\documentclass[manuscript, screen]{acmart}

\copyrightyear{2022}
\acmYear{2022}
\setcopyright{acmlicensed}\acmConference[To Appear]{CHI Conference on Human Factors in Computing Systems}{CHI 2022}{USA}
\acmBooktitle{CHI Conference on Human Factors in Computing Systems (CHI '22), April 29-May 5, 2022, New Orleans, LA, USA}
\acmPrice{15.00}
\acmDOI{10.1145/3491102.3501888}
\acmISBN{978-1-4503-9157-3/22/04}

\newcommand{\hl}[2]{{\color{#1}\bfseries [#2]}}
\usepackage{tabularx} 
\usepackage{booktabs} 
\usepackage[ruled]{algorithm2e} 
\usepackage[colorinlistoftodos]{todonotes}
\usepackage{verbatim}
\usepackage{graphicx}
\usepackage{xspace}

\usepackage{balance}       
\usepackage[T1]{fontenc}   
\usepackage{newtxmath}
\usepackage{hyperref}
\usepackage{color}
\usepackage{booktabs}

\usepackage{microtype}        
\usepackage{ccicons}          

\usepackage{cuted}
\usepackage{capt-of}

\usepackage{upquote}
\usepackage{listings}
\usepackage{xcolor}
\usepackage{float}

\newcommand\upquote[1]{\textquotesingle#1\textquotesingle}

\definecolor{codegreen}{rgb}{0,0.6,0}
\definecolor{codegray}{rgb}{0.5,0.5,0.5}
\definecolor{codepurple}{rgb}{0.58,0,0.82}

\lstdefinestyle{mystyle}{
    backgroundcolor=\color{white},
    commentstyle=\color{codegreen},
    keywordstyle=\color{magenta},
    numberstyle=\tiny\color{codegray},
    stringstyle=\color{codepurple},
    basicstyle=\ttfamily\footnotesize,
    breakatwhitespace=false,
    breaklines=true,
    captionpos=b,
    keepspaces=true,
    numbers=left,
    numbersep=5pt,
    showspaces=false,
    showstringspaces=false,
    showtabs=false,
    tabsize=2
}

\definecolor{light-gray}{gray}{0.95}

\definecolor{lightgray}{rgb}{0.83, 0.83, 0.83}

\usepackage{multirow}

\usepackage{blindtext}
\usepackage{subcaption}
\usepackage{caption}

\usepackage{enumitem}
\usepackage{graphicx}
\usepackage{multicol}

\usepackage{appendix}

\usepackage{tikz}
\usetikzlibrary{positioning,arrows.meta,graphs,backgrounds,fit,calc,quotes,shapes.multipart}
\usepackage{pgfplots}
\usepackage{pgfplotstable}

\pgfplotsset{width=3.5cm,
    tick label style={font=\footnotesize}
}

\definecolor{tisanecodetop}{RGB}{74,63,106}
\definecolor{closecolor}{RGB}{225,93,87}
\definecolor{minimizecolor}{RGB}{245,215,90}
\definecolor{maximizecolor}{RGB}{111,208,74}

\SetAlFnt{\small}
\SetAlCapFnt{\small}
\SetAlCapNameFnt{\small}
\SetAlCapHSkip{0pt}
\IncMargin{-\parindent}

\newcommand{\ej}[1]{\hl{purple}{EJ: #1}}
\newcommand{\as}[1]{\hl{blue}{AS: #1}}

\newcommand{\diff}[1]{\textcolor{black}{#1}}

\def\Disambiguation{Disambiguation\xspace}

\def\SDSLlong{study design specification language\xspace}
\def\SDSL{SDSL\xspace}

\def\dcConceptualKnowledge{\textit{DG1 - Conceptual knowledge}\xspace}
\def\dcConceptualKnowledgeLong{\textit{DG1 - Prioritize conceptual knowledge.}\xspace}
\def\dcValidity{\textit{DG2 - Validity}\xspace}
\def\dcValidityLong{\textit{DG2 - Prioritize the validity of models.}\xspace}
\def\dcGuidance{\textit{DG3 - Guidance and control}\xspace}
\def\dcGuidanceLong{\textit{DG3 - Give analysts guidance and control.}\xspace}
\def\dcStatisticalPlanning{\textit{DG4 - Statistical planning}\xspace}
\def\dcStatisticalPlanningLong{\textit{DG4 - Facilitate statistical planning without data.}\xspace}

\def\rqWorkflow{\textbf{RQ1 - Workflow}\xspace}
\def\rqCognitive{\textbf{RQ2 - Cognitive fixation}\xspace}
\def\rqFuture{\textbf{RQ3 - Future possibilities}\xspace}

\def\statsmodels{\texttt{statsmodels}\xspace}
\def\pymer{\texttt{pymer4}\xspace}

\usepackage{setspace}

\newcolumntype{H}{>{\setbox0=\hbox\bgroup}c<{\egroup}@{}}

\def\<#1>{\codeid{#1}}
\newcommand{\codeid}[1]{\ifmmode{\mbox{\small\ttfamily{#1}}}\else{\small\ttfamily #1}\fi}
\newcommand{\codeidsmall}[1]{\ifmmode{\mbox{\smaller\ttfamily{#1}}}\else{\smaller\ttfamily #1}\fi}

\def\plaintitle{Tisane: Authoring Statistical Models via Formal Reasoning from Conceptual and Data Relationships}
\def\plainauthor{Eunice Jun, Audrey Seo, Jeffrey Heer, Ren{\'e} Just}

\def\plainkeywords{statistical analysis; linear modeling; end-user programming; end-user elicitation; domain-specific language; transparent statistics; validity}

\makeatletter
\def\url@leostyle{%
  \@ifundefined{selectfont}{
    \def\UrlFont{\sf}
  }{
    \def\UrlFont{\small\bf\ttfamily}
  }}
\makeatother
\def\pprw{8.5in}
\def\pprh{11in}

\definecolor{linkColor}{RGB}{6,125,233}
\hypersetup{%
  pdftitle={\plaintitle},
 pdfauthor={\plainauthor},
  pdfkeywords={\plainkeywords},
  pdfdisplaydoctitle=true, 
  bookmarksnumbered,
  pdfstartview={FitH},
  colorlinks,
  citecolor=black,
  filecolor=black,
  linkcolor=black,
  urlcolor=linkColor,
  breaklinks=true,
  hypertexnames=false
}

\hypersetup{draft}

\begin{document}

\title{\plaintitle}

\author{Eunice Jun}
\email{emjun@cs.washington.edu}
\orcid{0000-0002-4050-4284}
\affiliation{%
  \institution{University of Washington}
  \city{Seattle}
  \state{Washington}
  \country{USA}
}
\author{Audrey Seo}
\email{alseo@cs.washington.edu }
\orcid{0000-0003-2928-3721}
\affiliation{%
  \institution{University of Washington}
  \city{Seattle}
  \state{Washington}
  \country{USA}
}
\author{Jeffrey Heer}
\email{jheer@cs.washington.edu}
\orcid{0000-0002-6175-1655}
\affiliation{%
  \institution{University of Washington}
  \city{Seattle}
  \state{Washington}
  \country{USA}
}
\author{Ren{\'e} Just}
\email{rjust@cs.washington.edu}
\orcid{0000-0002-5982-275X}
\affiliation{%
  \institution{University of Washington}
  \city{Seattle}
  \state{Washington}
  \country{USA}
}

\begin{abstract}

Proper statistical modeling incorporates domain theory about how concepts relate
and details of how data were measured. However, data analysts currently lack
tool support for recording and reasoning about domain assumptions, data
collection, and modeling choices in an integrated manner, leading to mistakes
that can compromise scientific validity. For instance, generalized linear
mixed-effects models (GLMMs) help answer complex research questions, but
omitting random effects impairs the generalizability of results. To address this
need, we present Tisane, a mixed-initiative system for authoring generalized linear models with and without mixed-effects.
Tisane introduces a \textit{study design specification language} for expressing and
asking questions about relationships between variables. Tisane contributes an
\textit{interactive compilation} process that represents relationships in a graph, infers
candidate statistical models, and asks follow-up questions to disambiguate user
queries to construct a valid model. In case studies with three researchers, we
find that Tisane helps them focus on their goals and assumptions while avoiding
past mistakes.
\end{abstract}

\maketitle






\def\myequation{$Y = \beta_{0} + \beta_{1} X_{1} + \cdots + \beta_{n} X_{n}$}

\newcommand{\figureSystemOverview}{
    \pgfplotstableread[col sep = comma]{figures/resid_plot.csv}\loadedtable

    \begin{figure}[h]
        \centering
        \begin{tikzpicture}[>=Stealth,
                            every node/.style={node distance=0.5cm}]
            \begin{scope}[local bounding box=bb,
                          every node/.style={inner sep=5pt, node distance=0.5cm},
                          stage/.style={fill=gray!10,rounded corners=4pt},
                          caption/.style={node distance=.2cm,rounded corners=4pt}]
                \node[stage] (input) at (0, 0) {\begin{tikzpicture}[every node/.style={inner sep=0pt},
                                                                    button/.style={circle,minimum size=3pt}]
                    \node[rectangle split, rectangle split parts=2, rounded corners=1pt, rectangle split part fill={tisanecodetop,white},rectangle split empty part height=7pt,inner sep=0pt,draw=gray!80,rectangle split draw splits=false] (n) {\nodepart{two} \begin{tikzpicture}
                        \node[inner sep=5pt,rectangle, minimum height=1.5cm]       {\footnotesize \texttt{import tisane as ts}};
                    \end{tikzpicture}};
                    \coordinate (start) at ($(n.text west)+(4pt,0)$);
                    \node[button,fill=closecolor]   (close) at (start) {};
                    \node[button,fill=minimizecolor,right=0.5pt of close] (minimize) {};
                    \node[button,fill=maximizecolor,right=0.5pt of minimize] (maximize) {};
                \end{tikzpicture}};
                \node[right=of input,stage] (graphir) {\begin{tikzpicture}[every node/.style={circle, draw=black,fill=white}]
                    \node (a) at (0,0) {};
                    \node (b) at (-0.5, -1) {};
                    \node (c) at (0.5, -1) {};
                    \graph{
                        (a) ->[densely dotted,bend right] (b);
                        (c) ->[bend right] (a);
                        (a) ->[bend right] (c);
                        (c) ->[densely dashed,bend left] (b);
                    };
                \end{tikzpicture}};
                \node[right=of graphir,stage] (disambig) {\includegraphics[width=.2\textwidth]{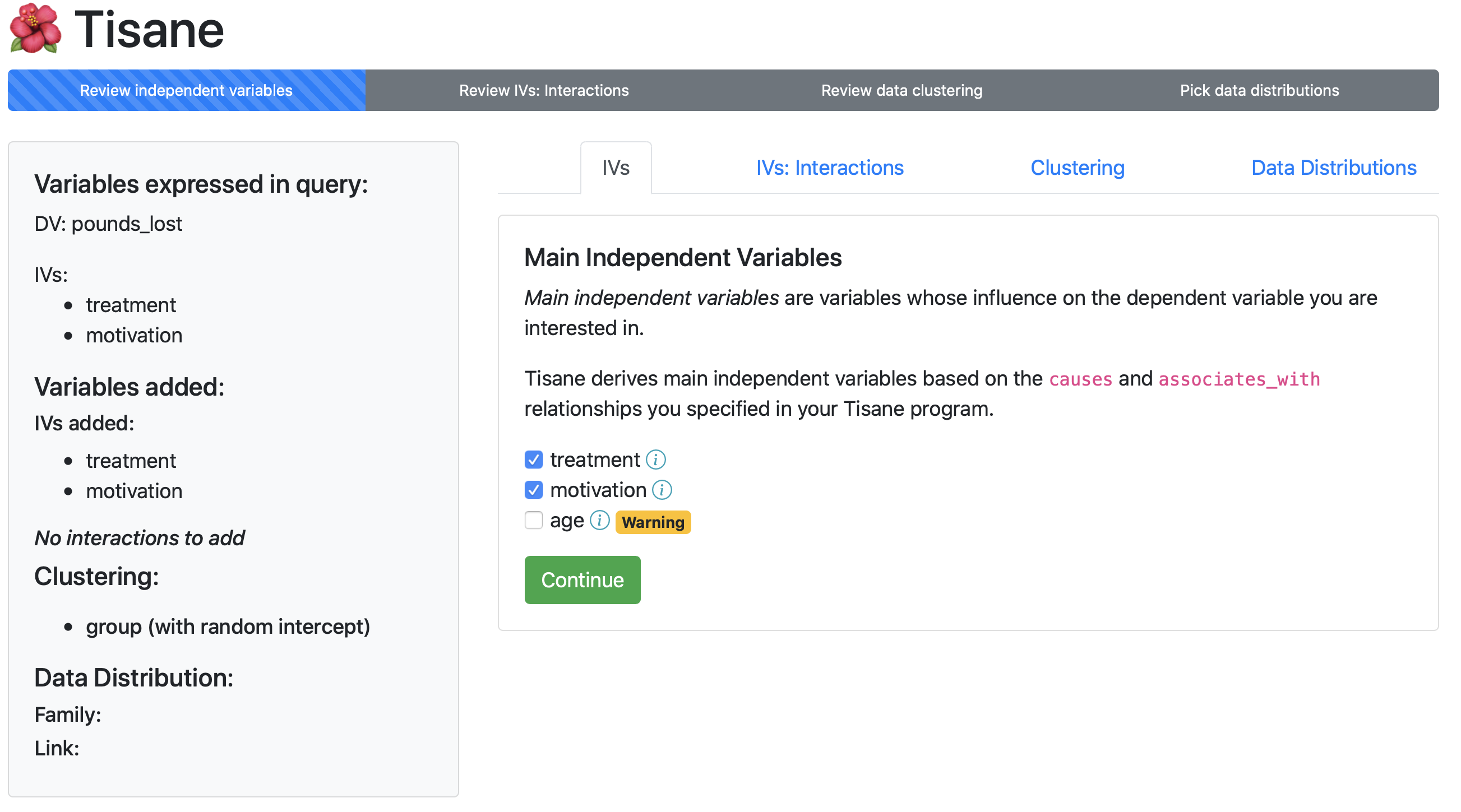}};
                \node[right=of disambig,stage,text width=.2\textwidth,align=center] (code) {{\footnotesize \myequation}\\%
                    \begin{tikzpicture}[every node/.style={rounded corners=0pt},rounded corners=0pt]
                        \begin{axis}[rounded corners=0pt,
                                     scatter/use mapped color={
                                        fill=blue!80,
                                        draw=blue!80
                                     },
                                     axis background/.style={fill=white},
                                     xmin=-6.9,
                                     xmax=6.9]
                             \draw[gray,very thin] (axis cs:-5,-0.5) -- (axis cs:-5,0.5);
                             \draw[gray,very thin] (axis cs:0,-0.5) -- (axis cs:0,0.5);
                             \draw[gray,very thin] (axis cs:5,-0.5) -- (axis cs:5,0.5);
                            \addplot[black,domain=-7:7] {0};
                            \addplot[scatter,only marks,mark size=1.1pt] table[x=x,y=y,col sep=comma] {\loadedtable};
                        \end{axis}
                        \pgfresetboundingbox
                        \path
                                ($(current axis.south west)-(0.2cm,0.35cm)$) rectangle ($(current axis.north east)+(0,0.2cm)$);
                    \end{tikzpicture}
                };
                \node   (codecaption) at ($(code.south)-(0,0.3cm)$)     {\textit{Output Code}};
                \coordinate (ref) at ($(codecaption) + (0.5cm,0)$) {};
                \node[caption]  (graphircaption1) at ($(codecaption)!(graphir)!(ref)$) {\textit{Graph IR}};
                \node[caption]  (disambigcaption1) at ($(codecaption)!(disambig)!(ref)$) {\textit{Disambiguation}};
                \node[caption]  (inputcaption)     at ($(codecaption)!(input)!(ref)$) {\textit{Input Study Design Specification}};
                \graph{
                    (input) ->[thick] (graphir) ->[thick] (disambig) ->[thick] (code);
                };
                \node at ($(current bounding box.south) + (0,-.4)$) {Interactive Compilation};
                \node[color=white,inner sep=0pt] at ($(current bounding box.north) + (0,5pt)$) {};
                \node[color=white,inner sep=0pt] (spacer) at ($(code.east)+(5pt,0)$) {};
            \end{scope}

            \begin{pgfonlayer}{background}
                \node[rounded corners=4pt, draw=gray!50, fit=(bb)] {};
            \end{pgfonlayer}
        \end{tikzpicture}
        \caption{\textbf{Overview of the Tisane system.} Analysts specify a set of variable relationships (\textit{Input Study Design Specification}). Tisane represents these in an internal graph (\textit{Graph IR}). To infer a statistical model, Tisane engages analysts in an interactive compilation process that elicits additional input from analysts in a disambiguation process (\textit{Disambiguation}) and outputs a script for fitting a valid GLM and visualizing its residuals (\textit{Output Code}).}
        \label{fig:figureSystemOverview}
        \Description{          Four boxes (which we will refer to as A, B, C, and D) are shown in a row. Box A is labeled “Input Study Design Specification” and shows a text editor window where the text “import tisane as ts” is written (this is the canonical way of importing Tisane in Python, like “import pandas as pd”). Box A has an arrow pointing to Box B, which is labeled “Graph IR,” where “IR” stands for “intermediate representation.” Box B contains a directed graph with three nodes, arranged like the vertices of an isosceles triangle. The top node has a dotted arrow pointing toward the bottom left node and a solid arrow pointing at the bottom right node. The bottom right node has a dashed edge pointing toward the bottom left node and a solid edge pointing to the top node. The bottom left node has no outgoing edges. This is meant to be a icon for the Graph IR. Box B has an arrow pointing to Box C, which is labeled as “Disambiguation.” Box C contains a screenshot of the Tisane GUI. The top of the GUI has the word “Tisane” and a progress bar, which is a quarter blue and three quarters gray. A left side panel gives an overview of the current model, and shows variables expressed by the query (the dependent variable, pounds_lost, and the independent variables, treatment and motivation), the variables added (the query independent variables are listed underneath, while noting that there are no interactions to add), clustering variables (group, a random intercept), and that for data distributions, no family or link functions have been selected. Box C has an arrow pointing to Box D, which is labeled “Output Code”. The box contains an equation, Y = beta_0 + beta_1 X_1 + … + beta_n X_n, and below the equation is a plot of residuals vs. fitted values. This is meant to represent the output script that Tisane creates.}
    \end{figure}
}

\newcommand{\figureGraphIRExample}{
    \begin{figure}[h]
        \begin{tikzpicture}[>=Stealth,
                            causes/.style={thick,draw=black, "causes", text=black},
                            associates/.style={thick,draw=black, "assoc."},
                            min/.style={minimum size=1.5cm},
                            unit/.style={min,draw=black},
                            measure/.style={min,circle,draw=black},
                            has/.style={densely dotted,thick},
                            nests/.style={dashed,thick},
                            depvar/.style={fill=gray!30},
                            every edge quotes/.style={fill=none,fill opacity=.9,text opacity=1,rounded corners=3pt,inner sep=2pt},
                            tightly/.style={inner sep=1pt},
                            loosely/.style={inner sep=.7em}]
            \node[unit]		(group) at (0,0)			{\texttt{group}};
            \node[unit,right=3.5cm of group]	(member)		{\texttt{member}};
            \node[measure,below=1.75cm of member]		(motivation)	{\texttt{motivation}};
            \coordinate (ref1) at ($(motivation.center) + (1cm, 0)$) {};
            \node[measure] at ($(ref1)!(group.center)!(motivation.center)$)		(condition)		{\texttt{condition}};

            \coordinate (midcoord1) at ($(group.center)!0.5!(member.center)$) {};
            \coordinate (midcoord2) at ($(member.center)!0.5!(motivation.center)$) {};
            \coordinate (ref) at ($(midcoord2) + (1cm, 0)$) {};
            \node[measure,depvar] at ($(midcoord2)!(midcoord1)!(ref)$)		(poundslost)	{\texttt{pounds\_lost}};

            \graph
                {
                (condition) ->[causes,sloped,below,pos=0.4,loosely] (poundslost);
                (motivation) ->[associates,bend right,pos=0.6,sloped,above] (poundslost);
                (poundslost)	->[associates,bend right,below,sloped]	(motivation);
                (member)		->[nests]		(group);
                (member)		->[has]			(poundslost);
                (member)		->[has]			(motivation);
                (group)			->[has]			(condition);
            };
        \end{tikzpicture}
        \caption{The graph representation of the variables and relationships from the usage scenario. \texttt{causes} edges are labeled with ``causes''. \texttt{associates\_with} edges are labeled with ``assoc.'' Dashed edges indicate \texttt{nests\_within} relationships, and dotted edges indicate \texttt{has} relationships.}
        \label{fig:figureGraphIRExample}
        \Description{A directed graph of five nodes is depicted. The nodes are labeled “group”, “condition”, “pounds_lost”, “motivation”, and “member”, all in monospace font. The “group” node is square-shaped (indicating the node represents a Unit), and has a dotted edge pointing to the circle-shaped (indicating it’s a Measure) “condition” node. The “condition” node has a solid edge labeled “causes” pointing to the circle-shaped, gray “pounds_lost” node. The gray color indicates that “pounds_lost” was the dependent variable. The “pounds_lost” node has one outgoing solid edge labeled “assoc.” (for associative relationships), to the circle-shaped node “motivation.” The “motivation” node has an outgoing, solid “assoc.” edges to the “pounds_lost” node. The final node, “member,” is square-shaped and has a dashed outgoing edge to the “group” node and dotted outgoing edges to the “pounds_lost” and “motivation” nodes. The dashed edge means “member” nests in “group” and the dotted edges mean that “pounds_lost” and “motivation” are measures of the “member” unit.}
    \end{figure}
}

\newcommand{\oldfigureCandidateMainEffects}{
    \begin{figure}[h]
        \def\arrowend{.4}
        \def\dotsbegin{.43}
        \def\dotsend{.55}
        \def\arrowbegin{.59}
        \begin{tikzpicture}
            \node[fill=gray!10,rounded corners=4pt]		(n1)			{%
                \begin{tikzpicture}[>=Stealth,
                                causes/.style={thick,draw=black, "causes",text=black},
                                associates/.style={thick,draw=black, "assoc."},
                                min/.style={minimum size=1cm},
                                unit/.style={min,draw=black},
                                measure/.style={fill=white,min,circle,draw=black},
                                has/.style={densely dotted,thick},
                                nests/.style={dashed,thick},
                                depvar/.style={fill=gray!30},
                                every edge quotes/.style={fill=none,fill opacity=.9,text opacity=1,rounded corners=3pt,inner sep=2pt},
                                tightly/.style={inner sep=1pt},
                                loosely/.style={inner sep=.7em},
                                scale=0.9,transform shape,
                                cme/.style={inner sep=0pt,text width=0.5cm,align=center,node font=\footnotesize}]
                \node[measure]	(iv1)		{IV};
                \node[measure,right=of iv1]	(iv2)		{IV};
                \node[measure,left=0.5cm of iv1]	(iv3)		{IV};
                \node[measure,above=of iv1]	(p1)		{CP};
                \node[measure,above=of iv2] (p2)		{CP};
                \node[measure,above=of iv3,cme]	(p3)		{CP\\CME};
                \coordinate	(point1) at ($ (p1) !.5! (p2) $)	{};
                \coordinate	(point2) at ($ (iv2) !.5! (iv3) $)	{};
                \node[measure] (a1)	at ($ (point1) + (0,2cm) $)	{CME};
                \draw (a1) edge[left,->] ($(a1)!\arrowend!(p1)$);
                \draw ($(a1)!\dotsbegin!(p1)$) edge[dotted,thick]	($(a1)!\dotsend!(p1)$);
                 edge[->] (p1);
                \draw ($(a1)!\arrowbegin!(p1)$) edge[left,->] (p1);

                \draw (a1) edge[right,->] ($(a1)!\arrowend!(p2)$);
                \draw ($(a1)!\dotsbegin!(p2)$) edge[dotted,thick]	($(a1)!\dotsend!(p2)$);
                 edge[->] (p1);
                \draw ($(a1)!\arrowbegin!(p2)$) edge[right,->] (p2);
                \graph{
                    (p1) -> (iv1);
                    (p2) -> (iv1);
                    (p2) -> (iv2);
                    (p3) -> (iv3);
                };
            \end{tikzpicture}%
            };

            \node[right=0.2cm of n1,fill=gray!10,rounded corners=4pt]	(n2)	{%
                \begin{tikzpicture}[>=Stealth,
                                    causes/.style={thick,draw=black, "causes",text=black},
                                    associates/.style={thick,draw=black, "assoc."},
                                    min/.style={minimum size=1cm},
                                    unit/.style={min,draw=black},
                                    measure/.style={fill=white,min,circle,draw=black},
                                    has/.style={densely dotted,thick},
                                    nests/.style={dashed,thick},
                                    depvar/.style={fill=gray!30},
                                    every edge quotes/.style={fill=none,fill opacity=.9,text opacity=1,rounded corners=3pt,inner sep=2pt},
                                    tightly/.style={inner sep=1pt},
                                    loosely/.style={inner sep=.7em},
                                    scale=0.9,transform shape,
                                    cme/.style={inner sep=0pt,text width=0.5cm,align=center,node font=\footnotesize}]
                    \node[measure]	(iv1)		{IV};
                    \node[measure,right=of iv1]	(iv2)		{IV};
                    \node[measure,right=of iv2]	(iv3)		{IV};
                    \coordinate	(point1) at ($ (iv1) !.5! (iv2) $)	{};
                    \coordinate	(point2) at ($ (iv2) !.5! (iv3) $)	{};
                    \node[measure,above=1.5cm of point1,cme]	(p1)		{SCA\\CME};
                    \node[measure,above=1.5cm of point2,cme] (p2)		{SCA\\CME};
                    \coordinate	(point3) at ($ (p1) !.5! (p2) $)	{};
                    \node[measure,above=1.5cm of point3,cme]	(p3)		{SCA\\CME};
                    \draw (p3) edge[->] ($(p3)!\arrowend!(p1)$);
                    \draw ($(p3)!0\dotsbegin!(p1)$) edge[dotted,thick]	($(p3)!\dotsend!(p1)$);
                    \draw ($(p3)!\arrowbegin!(p1)$) edge[left,->] (p1);

                    \draw (p3) edge[right,->] ($(p3)!\arrowend!(p2)$);
                    \draw ($(p3)!0\dotsbegin!(p2)$) edge[dotted,thick]	($(p3)!\dotsend!(p2)$);
                    \draw ($(p3)!\arrowbegin!(p2)$) edge[right,->] (p2);

                    \draw (p1) edge[->] ($(p1)!\arrowend!(iv1)$);
                    \draw ($(p1)!0\dotsbegin!(iv1)$) edge[dotted,thick]	($(p1)!\dotsend!(iv1)$);
                    \draw ($(p1)!\arrowbegin!(iv1)$) edge[->] (iv1);

                    \draw (p1) edge[->] ($(p1)!\arrowend!(iv2)$);
                    \draw ($(p1)!0\dotsbegin!(iv2)$) edge[dotted,thick]	($(p1)!\dotsend!(iv2)$);
                    \draw ($(p1)!\arrowbegin!(iv2)$) edge[->] (iv2);

                    \draw (p2) edge[->] ($(p2)!\arrowend!(iv2)$);
                    \draw ($(p2)!0\dotsbegin!(iv2)$) edge[dotted,thick]	($(p2)!\dotsend!(iv2)$);
                    \draw ($(p2)!\arrowbegin!(iv2)$) edge[->] (iv2);

                    \draw (p2) edge[->] ($(p2)!\arrowend!(iv3)$);
                    \draw ($(p2)!0\dotsbegin!(iv3)$) edge[dotted,thick]	($(p2)!\dotsend!(iv3)$);
                    \draw ($(p2)!\arrowbegin!(iv3)$) edge[->] (iv3);
                \end{tikzpicture}
            };
            \node[right=0.2cm of n2,fill=gray!10,rounded corners=4pt]	(n3)	{%
                \begin{tikzpicture}[>=Stealth,
                                    causes/.style={draw=black, "causes",text=black},
                                    associates/.style={draw=black, "assoc.",bend right},
                                    min/.style={minimum size=1cm},
                                    unit/.style={min,draw=black},
                                    measure/.style={fill=white,min,circle,draw=black},
                                    has/.style={densely dotted,thick},
                                    nests/.style={dashed,thick},
                                    depvar/.style={fill=gray!30},
                                    every edge quotes/.style={fill=none,fill opacity=.9,text opacity=1,rounded corners=3pt,inner sep=2pt,node font=\scriptsize},
                                    tightly/.style={inner sep=1pt},
                                    loosely/.style={inner sep=.7em},
                                    scale=0.9,transform shape]
                    \node[measure]	(iv)		{IV};
                    \node[measure,left=of iv]	(m1)		{\footnotesize CME};
                    \node[measure,right=of iv]	(m2)		{\footnotesize CME};
                    \node[measure,below=of iv]	(m3)		{};
                    \node[measure,depvar,above=of iv]	(dv)	{DV};

                    \graph{
                        (iv) ->[associates,below] (m2);
                        (iv) ->[associates,left] (m3);
                        (m2) ->[associates,above] (iv);
                        (m3) ->[associates,right] (iv);
                        (m1) ->[causes,above left] (dv);
                        (m2) ->[associates,above right] (dv);
                        (dv) ->[draw=black,"assoc.",below left,pos=0.25] (m2);
                    };
                \end{tikzpicture}
            };
            \coordinate	(ref1)	at	($(n2.south)-(1cm,0.3cm)$) {};
            \coordinate (ref2)	at	($(n2.south)-(-1cm,0.3cm)$) {};

            \node	(cap1)	at	($(ref1)!(n1)!(ref2)$)	{(a) \textit{oldest causal ancestors}};
            \node	(cap2)	at	($(ref1)!(n2)!(ref2)$)	{(b) \textit{shared causal ancestors}};
            \node[inner sep=0pt]	(cap3)	at	($(ref1)!(n3)!(ref2)$)	{(c) \textit{possible confounding variables}};
            \node[below=0.05cm of cap3,inner sep=0pt]	(cap4)	{\textit{\& possible causal omissions}};
        \end{tikzpicture}
        \caption{Graphs demonstrating oldest causal ancestors, shared causal ancestors, and possible confounding variables \& possible causal omissions. In graphs (a) and (b) (left and middle), all edges are causal. Independent variables are marked ``IV'', discovered candidate main effects ``CME'', dependent variables ``DV'', causal parents ``CP'', and shared causal ancestors ``SCA''.}
        \label{fig:figureCandidateMainEffects}
        \Description{Three directed graphs are shown, all with circle-shaped nodes, labeled “(a) causal parents”, “(b) possible causal omissions”, and “(c) possible confounding associations”. Graph (a) contains six nodes, arranged in three columns of two nodes each. In each column, the node on the top is labeled “CP CME” and the node on the bottom is labeled “IV”. The top left node (“CP CME”), has an edge to the node “IV” below it. The top middle node (“CP CME”) has an edge to the node “IV” below it. The top right node (“CP CME”) has two outgoing edges, to the bottom middle node (“IV”) and to the node “IV” below it. Graph (b) contains two nodes. One node is labeled “CME”, and has an edge pointing to a gray node labeled “DV”. Graph (c) contains four nodes. One node is unlabeled, and the rest are labeled “IV”, “CME”, and “DV”. The “DV” node is gray colored. There are associative edges between the unlabeled node and “IV”, between “IV” and “CME”, and between “CME” and “DV.”}
    \end{figure}
}
\newcommand\betweendistance[1]{1.5cm of #1}
\newcommand{\figureCandidateMainEffects}{
    \begin{figure}[h]
        \def\arrowend{.4}
        \def\dotsbegin{.43}
        \def\dotsend{.55}
        \def\arrowbegin{.59}

        \begin{tikzpicture}
            \node[fill=gray!10,rounded corners=4pt]		(n1)			{%
                \begin{tikzpicture}[>=Stealth,
                                causes/.style={thick,draw=black, "causes",text=black},
                                associates/.style={thick,draw=black, "assoc."},
                                min/.style={minimum size=1cm},
                                unit/.style={min,draw=black},
                                measure/.style={fill=white,min,circle,draw=black},
                                has/.style={densely dotted,thick},
                                nests/.style={dashed,thick},
                                depvar/.style={fill=gray!30},
                                every edge quotes/.style={fill=none,fill opacity=.9,text opacity=1,rounded corners=3pt,inner sep=2pt},
                                tightly/.style={inner sep=1pt},
                                loosely/.style={inner sep=.7em},
                                scale=0.9,transform shape,
                                cme/.style={inner sep=0pt,text width=0.5cm,align=center,node font=\footnotesize}]
                \node[measure]	(iv1)		{IV};
                \node[measure,right=of iv1]	(iv2)		{IV};
                \node[measure,left=0.5cm of iv1]	(iv3)		{IV};
                \node[measure,above=of iv1,cme]	(p1)		{CP\\CME};
                \node[measure,above=of iv2,cme] (p2)		{CP\\CME};
                \node[measure,above=of iv3,cme]	(p3)		{CP\\CME};
                \coordinate	(point1) at ($ (p1) !.5! (p2) $)	{};
                \coordinate	(point2) at ($ (iv2) !.5! (iv3) $)	{};

                \graph{
                    (p1) -> (iv1);
                    (p2) -> (iv1);
                    (p2) -> (iv2);
                    (p3) -> (iv3);
                };
            \end{tikzpicture}%
            };

            \node[right=1.5cm of n1,fill=gray!10,rounded corners=4pt]	(n2)	{%
                \begin{tikzpicture}[>=Stealth,
                                    causes/.style={thick,draw=black, "causes",text=black},
                                    associates/.style={thick,draw=black, "assoc."},
                                    min/.style={minimum size=1cm},
                                    unit/.style={min,draw=black},
                                    measure/.style={fill=white,min,circle,draw=black},
                                    has/.style={densely dotted,thick},
                                    nests/.style={dashed,thick},
                                    depvar/.style={fill=gray!30},
                                    every edge quotes/.style={fill=none,fill opacity=.9,text opacity=1,rounded corners=3pt,inner sep=2pt},
                                    tightly/.style={inner sep=1pt},
                                    loosely/.style={inner sep=.7em},
                                    scale=0.9,transform shape,
                                    cme/.style={inner sep=0pt,text width=0.5cm,align=center,node font=\footnotesize}]
                    \node[measure]  (cme)   {CME};
                    \node[measure,depvar,below=of cme]  (dv)    {DV};
                    \graph{
                        (cme) -> (dv);
                    };
                \end{tikzpicture}
            };
            \node[right=1.5cm of n2,fill=gray!10,rounded corners=4pt]	(n3)	{%
                \begin{tikzpicture}[>=Stealth,
                                    causes/.style={draw=black, "causes",text=black},
                                    associates/.style={draw=black, "assoc.",bend right},
                                    min/.style={minimum size=1cm},
                                    unit/.style={min,draw=black},
                                    measure/.style={fill=white,min,circle,draw=black},
                                    has/.style={densely dotted,thick},
                                    nests/.style={dashed,thick},
                                    depvar/.style={fill=gray!30},
                                    every edge quotes/.style={fill=none,fill opacity=.9,text opacity=1,rounded corners=3pt,inner sep=2pt,node font=\scriptsize},
                                    tightly/.style={inner sep=1pt},
                                    loosely/.style={inner sep=.7em},
                                    scale=0.9,transform shape,
                                    smalldot/.style={circle,minimum size=5pt,fill=blue,draw=none}]
                    \node[measure]	(iv)		{IV};
                    \node[measure,depvar,above=of iv]	(dv)	{DV};
                    \coordinate (center)    at ($(iv)!0.5!(dv)$);
                    \coordinate (centeroffset) at ($(center)+ (3pt,0)$);
                    \coordinate[right=of iv.east] (rightcoord);
                    \coordinate[left=of iv.west] (leftcoord);
                    \coordinate (leftleftcoord) at ($(leftcoord)-(1cm,0)$);

                    \node[measure]	(m2)	at ($(center)!(rightcoord)!(centeroffset)$)	{\footnotesize CME};
                    \node[measure]	(m3)    at ($(centeroffset)!(leftleftcoord)!(center)$)		{};

                    \graph{
                        (iv) ->[associates,above left] (m2);
                        (iv) ->[associates,below left] (m3);
                        (m2) ->[associates,below right] (iv);
                        (m3) ->[associates,above right] (iv);
                        (m2) ->[associates,below left] (dv);
                        (dv) ->[associates,above right] (m2);
                    };
                \end{tikzpicture}%
            };
            \coordinate	(ref1)	at	($(n2.south)-(1cm,0.3cm)$) {};
            \coordinate (ref2)	at	($(n2.south)-(-1cm,0.3cm)$) {};

            \node	(cap1)	at	($(ref1)!(n1)!(ref2)$)	{(a) \textit{causal parents}};
            \node	(cap2)	at	($(ref1)!(n2)!(ref2)$)	{(b) \textit{possible causal omissions}};
            \node[inner sep=0pt]	(cap3)	at	($(ref1)!(n3)!(ref2)$)	{(c) \textit{possible confounding associations}};
        \end{tikzpicture}
        \caption{Graphs demonstrating causal parents, possible causal omissions, and possible confounding associations. In graphs (a) and (b) (left and middle), all edges are causal. Independent variables are marked ``IV'', discovered candidate main effects ``CME'', dependent variables ``DV'', and causal parents ``CP''.}
        \label{fig:figureCandidateMainEffects}
    \end{figure}
}

\newcommand{\figureOnlyAssociatesOrCausesEdgesCandidateMainEffects}{
    \begin{figure}[H]
        \def\arrowend{.4}
        \def\dotsbegin{.43}
        \def\dotsend{.55}
        \def\arrowbegin{.59}
        \begin{tikzpicture}
            \node[fill=gray!10,rounded corners=4pt]	(n3)	{%
                \begin{tikzpicture}[>=Stealth,
                                    causes/.style={draw=black, "causes",text=black},
                                    associates/.style={draw=black, "assoc.",bend right},
                                    min/.style={minimum size=1cm},
                                    unit/.style={min,draw=black},
                                    measure/.style={fill=white,min,circle,draw=black},
                                    has/.style={densely dotted,thick},
                                    nests/.style={dashed,thick},
                                    depvar/.style={fill=gray!30},
                                    every edge quotes/.style={fill=none,fill opacity=.9,text opacity=1,rounded corners=3pt,inner sep=2pt,node font=\scriptsize},
                                    tightly/.style={inner sep=1pt},
                                    loosely/.style={inner sep=.7em},
                                    scale=0.9,transform shape]
                    \node[measure]	(iv)		{IV};
                    \node[measure,left=of iv]	(m1)		{\footnotesize CME};
                    \node[measure,right=of iv]	(m2)		{\footnotesize CME};
                    \node[measure,below=of iv]	(m3)		{};
                    \node[measure,depvar,above=of iv]	(dv)	{DV};

                    \graph{
                        (iv) ->[associates,below] (m2);
                        (iv) ->[associates,left] (m3);
                        (m2) ->[associates,above] (iv);
                        (m3) ->[associates,right] (iv);
                        (dv) ->[associates,below right] (m1);
                        (m1) ->[associates,above left] (dv);
                        (m2) ->[associates,above right] (dv);
                        (dv) ->[draw=black,"assoc.",below left,pos=0.25] (m2);
                    };
                \end{tikzpicture}
            };

        \end{tikzpicture}
        \caption{A graph demonstrating an edge case for candidate main effect identification, where the graph contains only associative edges. Candidate main effects are labeled ``CME'', independent variables ``IV'', and dependent variables ``DV''. Variables that are none of the above are left unlabeled. When a graph contains only associative edges, candidate main effects are identified as those that are either associated with the DV or are associated with both the IV and the DV. (Note that the graph could contain additional edges/nodes other than the ones pictured, but the additional edges would not violate any of the initial checks that Tisane makes on the graph IR.)}
        \label{fig:figureOnlyAssociatesOrCausesEdgesCandidateMainEffects}
    \end{figure}
}

\newcommand{\figureSDSLToGraphIR}{
    \def\minsize{0.8cm}
    \def\mynodefont{\normalsize}
    \def\myedgequotesfont{\scriptsize}
    \def\mynodedistance{0.9cm}
    \begin{figure}[H]
        \begin{tikzpicture}[minigraph/.style={fill=gray!10,rounded corners=4pt},
                            caption/.style={node distance=.2cm,rounded corners=4pt,node font=\footnotesize},
                            closecaption/.style={node distance=.05cm,rounded corners=4pt,node font=\footnotesize}]
            \node[minigraph]		(causes)			{%
                \begin{tikzpicture}[>=Stealth,
                                every node/.style={node distance=\mynodedistance},
                                causes/.style={draw=black, "causes",text=black},
                                associates/.style={draw=black, "assoc."},
                                min/.style={minimum size=\minsize,node font=\mynodefont},
                                unit/.style={min,fill=white,draw=black},
                                measure/.style={fill=white,min,circle,draw=black},
                                has/.style={densely dotted,thick,"has"},
                                nests/.style={dashed,thick,"nests"},
                                depvar/.style={fill=gray!30},
                                every edge quotes/.style={node font=\myedgequotesfont,fill=none,fill opacity=.9,text opacity=1,rounded corners=3pt,inner sep=2pt}]
                    \node[measure]       (v1)        {\texttt{v1}};
                    \node[measure,right=of v1]  (v2) {\texttt{v2}};
                    \graph{
                        (v1) ->[causes,below] (v2);
                    };
                \end{tikzpicture}
            };

            \node[minigraph,below=of causes]		(assoc)			{%
                \begin{tikzpicture}[>=Stealth,
                                every node/.style={node distance=\mynodedistance},
                                causes/.style={draw=black, "causes",text=black},
                                associates/.style={draw=black, "assoc.",bend right},
                                min/.style={minimum size=\minsize,node font=\mynodefont},
                                unit/.style={min,fill=white,draw=black},
                                measure/.style={fill=white,min,circle,draw=black},
                                has/.style={densely dotted,thick,"has"},
                                nests/.style={dashed,thick,"nests"},
                                depvar/.style={fill=gray!30},
                                every edge quotes/.style={node font=\myedgequotesfont,fill=none,fill opacity=.9,text opacity=1,rounded corners=3pt,inner sep=2pt}]
                    \node[measure]       (v1)        {\texttt{v1}};
                    \node[measure,right=of v1]  (v2) {\texttt{v2}};
                    \graph{
                        (v1) ->[associates,above] (v2);
                        (v2) ->[associates,below] (v1);
                    };
                \end{tikzpicture}
            };
            \coordinate (mid)   at ($(causes.east)!0.5!(assoc.east)$);

            \node[minigraph,right=of mid]		(moderates)			{%
                \begin{tikzpicture}[>=Stealth,
                                every node/.style={node distance=\mynodedistance},
                                causes/.style={draw=black, "causes",text=black},
                                associates/.style={draw=black, "assoc.",bend right},
                                min/.style={minimum size=\minsize,node font=\mynodefont},
                                unit/.style={min,fill=white,draw=black},
                                measure/.style={fill=white,min,circle,draw=black},
                                has/.style={densely dotted,"has"},
                                nests/.style={dashed,thick,"nests"},
                                depvar/.style={fill=gray!30},
                                every edge quotes/.style={node font=\myedgequotesfont,fill=none,fill opacity=.9,text opacity=1,rounded corners=3pt,inner sep=2pt}]
                    \node[measure]       (v3)        {\texttt{m3}};
                    \node[measure,left=of v3,inner sep=0pt]  (v1v2) {\footnotesize \texttt{m1*m2}};
                    \node[measure,above right=of v3]    (v1)          {\texttt{m1}};
                    \node[measure,below right=of v3]    (v2)          {\texttt{m2}};
                    \coordinate (ref) at ($(v2) + (3pt,0)$);
                    \graph{
                        (v1v2) ->[associates,below] (v3);
                        (v3) ->[associates,above] (v1v2);
                        (v1) ->[associates,above left] (v3);
                        (v3) ->[associates,below right] (v1);
                        (v2) ->[associates,above right] (v3);
                        (v3) ->[associates,below left] (v2);
                    };
                \end{tikzpicture}
            };

            \node[minigraph,inner sep=0pt,right=of moderates]		(nests)			{%
                \begin{tikzpicture}[>=Stealth,
                                every node/.style={node distance=\mynodedistance},
                                causes/.style={draw=black, "causes",text=black},
                                associates/.style={draw=black, "assoc.",bend right},
                                min/.style={minimum size=\minsize,node font=\mynodefont},
                                unit/.style={min,fill=white,draw=black,rounded corners=0pt},
                                measure/.style={fill=white,min,circle,draw=black},
                                has/.style={densely dotted,thick,"has"},
                                nests/.style={dashed,"nests"},
                                depvar/.style={fill=gray!30},
                                every edge quotes/.style={node font=\myedgequotesfont,fill=none,fill opacity=.9,text opacity=1,rounded corners=3pt}]
                    \node[unit]       (u1)        {\texttt{u1}};
                    \node[unit,above=of u1]  (u2) {\texttt{u2}};
                    \node[inner sep=0pt] at ($(u2.north)+(0,3pt)$)  {};
                    \node[inner sep=0pt] at ($(u2.east)+(2pt,0)$) {};
                    \node[inner sep=0pt] at ($(u2.west)-(3pt,0)$) {};
                    \node[inner sep=0pt] at ($(u1.south)-(0,3pt)$) {};
                    \coordinate (ref)   at ($(u1)!0.5!(u2)$);
                    \node[right=2pt of ref,inner sep=0pt,node font=\scriptsize] {nests};
                    \graph{
                        (u1) ->[dashed] (u2);
                    };
                \end{tikzpicture}};

            \node[minigraph,right=of nests,inner sep=0pt]		(has)			{%
                \begin{tikzpicture}[>=Stealth,
                                every node/.style={node distance=\mynodedistance},
                                causes/.style={draw=black, "causes",text=black},
                                associates/.style={draw=black, "assoc.",bend right},
                                min/.style={minimum size=\minsize,node font=\mynodefont},
                                unit/.style={min,fill=white,draw=black,rounded corners=0pt},
                                measure/.style={fill=white,min,circle,draw=black},
                                has/.style={densely dotted,"has"},
                                nests/.style={dashed,"nests"},
                                depvar/.style={fill=gray!30},
                                every edge quotes/.style={node font=\myedgequotesfont,fill=none,fill opacity=.9,text opacity=1,rounded corners=3pt,inner sep=2pt}]
                    \node[unit]       (u)        {\texttt{u}};
                    \node[measure,below=of u]  (m) {\texttt{m}};
                    \node[inner sep=0pt] at ($(u.north)+(0,3pt)$)  {};
                    \node[inner sep=0pt] at ($(u.east)+(2pt,0)$) {};
                    \node[inner sep=0pt] at ($(u.west)-(3pt,0)$) {};
                    \node[inner sep=0pt] at ($(m.south)-(0,3pt)$) {};
                    \graph{
                        (u) ->[has,right] (m);
                    };
                \end{tikzpicture}
            };

            \node[caption,below=of causes]  (causescaption) {\texttt{v1.causes(v2)}};
            \node[caption,below=of assoc]   (assoccaption)  {\texttt{v1.associates\_with(v2)}};
            \node[caption,below=of moderates]   (moderatescaption)  {\texttt{m1.moderates(m2, on=m3)}};
            \node[caption,below=of nests]       (nestscaption)      {\texttt{u1.nests\_within(u2)}};
            \node[caption,below=of has]         (hascaption)        {\texttt{u.has(m)}};
            \node[closecaption,left=of causes]   {(a)};
            \node[closecaption,left=of assoc]    {(b)};
            \node[closecaption,left=of moderates]    {(c)};
            \node[closecaption,left=of nests]    {(d)};
            \node[closecaption,left=of has]      {(e)};

        \end{tikzpicture}
        \caption{Code snippets of conceptual and data measurement relationships written in Tisane's \SDSLlong and their representation in Tisane's graph IR. Variables are named with \texttt{u} for units, \texttt{m} for measures, and \texttt{v} for data variables that can be either units or measures. All edges depicted are those that are added due to the relationship. In the \texttt{moderates} example, we assume that \texttt{m1} and \texttt{m2} both belong to the same unit, and for simplicity, the attribution edge (labeled as ``has'') from \texttt{m1} and \texttt{m2}'s unit is not shown. For more complex examples, see the supplemental materials.}
        \label{fig:figureSDSLToGraphIR}
        \Description{Five directed graphs are shown. All nodes are labeled with a monospaced font. The first directed graph has two nodes, labeled “v1” and “v2”, both circles. There is a solid edge from “v1” to “v2” labeled “causes”. Below this graph is a code snippet reading “v1.causes(v2)”. The second directed graph also has two nodes, also labeled “v1” and “v2” and both circles. There are solid edges from “v1” to “v2” and from “v2” to “v1”. Both edges are labeled “assoc.” Below the graph is a code snippet: “v1.associates_with(v2)”. The third directed graph contains four nodes, which are labeled “m1*m2”, “m1”, “m2”, and “m3”. All are circles. There are six edges in the graph, all solid and labeled “assoc.” The edges are from “m1*m2” to “m3”, “m3” to “m1*m2”, “m1” to “m3”, “m3” to “m1”, “m2” to “m3”, and “m3” to “m2”. Below the graph is a code snippet: “m1.moderates(m2, on=m3)”. The fourth directed graph contains two nodes, labeled “u1” and “u2”. Both are squares. There is a dashed edge labeled “nests” from “u1” to “u2”. Below the graph is the code snippet “u1.nests_within(u2)”. The fifth, and final, directed graph contains two nodes labeled “u” and “m”. “u” is square-shaped and “m” is circle-shaped. There is a dotted edge from “u” to “m”, labeled “has”. Below the graph is the code snippet “u.has(m)”.}
    \end{figure}
}

\newcommand{\figureComplexModeratesGraphIR}{
    \def\minsize{0.8cm}
    \def\mynodefont{\normalsize}
    \def\myedgequotesfont{\scriptsize}
    \def\mynodedistance{0.9cm}
    \def\myx{.5cm}
    \def\myy{.9cm}
    \def\mone{\texttt{m1}\xspace}
    \def\mtwo{\texttt{m2}\xspace}
    \def\mthree{\texttt{m3}\xspace}
    \def\uone{\texttt{u1}\xspace}
    \def\utwo{\texttt{u2}\xspace}
    \def\u{\texttt{u}\xspace}
    \def\umone{\texttt{u1*m1}\xspace}
    \def\monetwo{\texttt{m1*m2}\xspace}
    \begin{figure}[H]
        \begin{tikzpicture}[minigraph/.style={fill=gray!10,rounded corners=4pt, minimum height=3.7cm},
                            caption/.style={node distance=.2cm,rounded corners=4pt,node font=\footnotesize}]
            \node[minigraph]		(moderates1)			{%
                \begin{tikzpicture}[>=Stealth,
                                every node/.style={node distance=\mynodedistance,minimum height=0cm},
                                causes/.style={draw=black, "causes",text=black},
                                associates/.style={draw=black, "assoc.",bend right},
                                min/.style={minimum size=\minsize,node font=\mynodefont},
                                unit/.style={min,fill=white,draw=black,rounded corners=0pt},
                                measure/.style={fill=white,min,circle,draw=black},
                                has/.style={densely dotted,"has"},
                                nests/.style={dashed,thick,"nests"},
                                depvar/.style={fill=gray!30},
                                every edge quotes/.style={node font=\myedgequotesfont,fill=none,fill opacity=.9,text opacity=1,rounded corners=3pt,inner sep=2pt}]
                    \node[measure]       (m2)        {\texttt{m2}};
                    \node[measure,left=of m2,inner sep=0pt]  (um1) {\footnotesize \texttt{u1*m1}};
                    \node[unit,above right=of m2]    (u1)          {\texttt{u1}};
                    \node[measure,below right=of m2]    (m1)          {\texttt{m1}};
                    \coordinate (ref) at ($(m1) + (3pt,0)$);
                    \node[unit] (u2) at ($(m1)!(um1)!(ref)$)         {\texttt{u2}};
                    \graph{
                        (um1) ->[associates,below] (m2);
                        (m2) ->[associates,above] (um1);
                        (u1) ->[associates,above left] (m2);
                        (m2) ->[associates,below right] (u1);
                        (m1) ->[associates,above right] (m2);
                        (m2) ->[associates,below left] (m1);
                        (u2)  ->[has,below,draw=gray,text=gray] (m1);
                        (u2)  ->[has,right,near start] (um1);
                    };
                \end{tikzpicture}
            };
            \node[minigraph, right=of moderates1]		(moderates2)			{%
                \begin{tikzpicture}[>=Stealth,
                                every node/.style={node distance=\mynodedistance,minimum height=0cm},
                                causes/.style={draw=black, "causes",text=black},
                                associates/.style={draw=black, "assoc.",bend right},
                                min/.style={minimum size=\minsize,node font=\mynodefont},
                                unit/.style={min,fill=white,draw=black,rounded corners=0pt},
                                measure/.style={fill=white,min,circle,draw=black},
                                has/.style={densely dotted,"has"},
                                old/.style={draw=gray,text=gray},
                                nests/.style={dashed,thick,"nests"},
                                depvar/.style={fill=gray!30},
                                every edge quotes/.style={node font=\myedgequotesfont,fill=none,fill opacity=.9,text opacity=1,rounded corners=3pt,inner sep=2pt}]
                    \node[measure]       (v3)        {\texttt{m3}};
                    \node[measure,left=of v3,inner sep=0pt]  (v1v2) {\footnotesize \texttt{m1*m2}};
                    \node[measure,above right=of v3]    (v1)          {\texttt{m1}};
                    \node[measure,below right=of v3]    (v2)          {\texttt{m2}};
                    \coordinate (ref) at ($(v2) + (3pt,0)$);
                    \coordinate (ref1) at ($(v1) + (3pt,0)$);
                    \node[unit] (u) at ($(v2)!(v1v2.west)!(ref)$)         {\texttt{u2}};
                    \node[unit] (u2) at ($(v1)!(v1v2.west)!(ref1)$)       {\texttt{u1}};
                    \graph{
                        (v1v2) ->[associates,below] (v3);
                        (v3) ->[associates,above] (v1v2);
                        (v1) ->[associates,above left] (v3);
                        (v3) ->[associates,below right] (v1);
                        (v2) ->[associates,above right] (v3);
                        (v3) ->[associates,below left] (v2);
                        (u)  ->[has,below,old] (v2);
                        (u)  ->[has,right,near start] (v1v2);
                        (u2) ->[has,above,old] (v1);
                        (u2) ->[has,right,near start] (v1v2);
                    };
                \end{tikzpicture}
            };

            \node[minigraph, right=of moderates2]		(moderates3)			{%
                \begin{tikzpicture}[>=Stealth,
                                every node/.style={node distance=\mynodedistance,minimum height=0cm},
                                causes/.style={draw=black, "causes",text=black},
                                associates/.style={draw=black, "assoc.",bend right},
                                min/.style={minimum size=\minsize,node font=\mynodefont},
                                unit/.style={min,fill=white,draw=black,rounded corners=0pt},
                                measure/.style={fill=white,min,circle,draw=black},
                                has/.style={densely dotted,"has"},
                                old/.style={draw=gray,text=gray},
                                nests/.style={dashed,thick,"nests"},
                                depvar/.style={fill=gray!30},
                                every edge quotes/.style={node font=\myedgequotesfont,fill=none,fill opacity=.9,text opacity=1,rounded corners=3pt,inner sep=2pt}]
                    \node[measure]       (v3)        {\texttt{m3}};
                    \node[measure,left=2cm of v3,inner sep=0pt]  (v1v2) {\footnotesize \texttt{m1*m2}};
                    \node[measure,above right=1.5cm and 0.6cm of v3]    (v1)          {\texttt{m1}};
                    \node[measure,above left=0.9cm and 0.5cm of v3]    (v2)          {\texttt{m2}};
                    \coordinate (ref) at ($(v2) + (3pt,0)$);
                    \coordinate (ref1) at ($(v1) + (3pt,0)$);
                    \node[unit] (u2) at ($(v1)!(v1v2.west)!(ref1)$)       {\texttt{u}};
                    \graph{
                        (v1v2) ->[associates,above] (v3);
                        (v3) ->[associates,below] (v1v2);
                        (v1) ->[associates,sloped,below] (v3);
                        (v3) ->[associates,sloped,above] (v1);
                        (v2) ->[associates,sloped,above] (v3);
                        (v3) ->[associates,sloped,below] (v2);
                        (u2) ->[has,above,old] (v1);
                        (u2) ->[has,below left,old] (v2);
                        (u2) ->[has,right] (v1v2);
                    };
                \end{tikzpicture}
            };
            \node[caption,below=of moderates1]      {(a) \texttt{m1.moderates(u1, on=m2)}};
            \node[caption,below=of moderates2]      {(b) \texttt{m1.moderates(m2, on=m3)}};
            \node[caption,below=of moderates3]      {(c) \texttt{m1.moderates(m2, on=m3)}};
        \end{tikzpicture}
        \caption{More complex examples of \texttt{moderates} written in Tisane's \SDSLlong, and their representation in Tisane's graph IR. Variables are named with \texttt{u} for units, \texttt{m} for measures, and \texttt{v} for data variables that can be either units or measures. Black edges have been added due to the \texttt{moderates} relationship. Gray edges already existed in the graph. In (a), only \mone is a measure, whose unit is \utwo, so \umone inherits an attribution edge only from \utwo. In (b), \mone and \mtwo are measures, with units \uone and \utwo respectively, so \monetwo inherits attribution edges from both \uone and \utwo. In (c), measures \mone and \mtwo share a unit, \u, and \monetwo inherits only one attribution edge from \u.}
        \label{fig:figureComplexModeratesGraphIR}
    \end{figure}
}

\newcommand{\tableClusteringExamples}{
    \begin{table}
        \caption{\textbf{Common types of data clustering that Tisane
        automatically accounts for inferred statistical models.} \ej{Should the
        examples come from the preliminary expressive coverage analysis?} There
        are three types of clustering that Tisane detects based on analysts'
        data measurement relationships. Below are examples for each type of
        clustering and the maximal random effects~\cite{barr2013random} Tisane
        derives. Tisane also derives random effects for interaction terms based
        on Barr's updated
        rules~\cite{barr2013randomUpdated}.}\label{table:clusteringExamples}
        \begin{tabular}{>{\raggedright}p{0.2\linewidth}>{\raggedright}p{0.5\linewidth}>{\raggedright\arraybackslash}p{0.3\linewidth}}
            Type of clustering	&	Example	&	Random effects Tisane infers \\
            \hline
            Repeated measures	&	Typing speed is measured once per day for five days. 	&	Random intercept for individuals, random intercept for days \\
            Hierarchical data 	&	Adults in exercise groups participate in a study where each group receives a different training regimen.~\cite{cohen2013applied}	&	Random intercept for group \\
            Non-nesting composition 	&	{Participants are assigned two conditions, each representing one of the five senses. In each condition, participants use an input device that leverages a different sense. Each input device is designed for exactly one sense, so input devices are not independent of condition.}	&	Random intercept and slope for participants, random intercept for input devices
        \end{tabular}
        \label{tab:tableClusteringExamples}
    \end{table}
}

\newcommand{\tableFamilyLinkFunctions}{
  \begin{table}[h]
    \caption{The available family and link functions in Tisane. Tisane generates code to fit models using \texttt{statsmodels} and \texttt{pymer4}. The package \texttt{statsmodels} supports GLMs without mixed-effects and a wider variety of family and link function combinations. The package \texttt{pymer4} supports GLMs with mixed effects and has much more limited support for family and link functions. As \texttt{statsmodels} and \texttt{pymer} add more support, Tisane can be extended.}
    \begin{tabular}{lll} \hline
      &	\multicolumn{2}{c}{Link functions (*default)} \\ \cline{2-3}
      \multicolumn{1}{c}{Family functions} & \multicolumn{1}{>{\raggedright}p{0.35\linewidth}}{Generalized linear models without mixed effects (\texttt{statsmodels})} & \multicolumn{1}{>{\raggedright}p{0.35\linewidth}}{Generalized linear models with mixed effects (\texttt{pymer4})}	\\ \hline
      \multirow{1}{*}{Gaussian} 	&	\multicolumn{1}{>{\raggedright}p{0.35\linewidth}}{Identity*, Inverse, Log}	&	\multirow{1}{*}{Identity*}	\\
      Inverse Gaussian	&	Identity, Inverse, Inverse Squared*, Log	&	Inverse Squared*	\\
      Gamma	&	Identity, Inverse*, Log 	&	Inverse*	\\
      Poisson 	&	Identity, Log*, Square Root	&	Log*	\\
      Binomial	&	Cauchy, CLogLog, Log, Logit*, Probit,	&	Logit*	\\
      Negative Binomial	&	Identity, Log*, Logit, Probit	&	N/A	\\
      Tweedie Family	&	Log*, Power	& N/A		\\
    \end{tabular}
    \label{tab:tableFamilyLinkFunctions}
  \end{table}
}

\newcommand{\michaelsFirstModel}{
    \lstinputlisting[
        language=Python,
        caption={Michael's first model attempt, from the usage scenario (\autoref{sec:usage_scenario}). Michael specifies \poundslost as his dependent variable, and \regimen and \motivation as his independent variables.},
        linerange={1-8},
        label={lst:michaelsFirstModel}
    ]{code/usage_scenario_group_exercise_statsmodels.py}
}

\newcommand{\michaelsSecondModel}{
    \lstinputlisting[
        language=Python,
        caption={Michael's second model attempt. Building on his first model (\autoref{lst:michaelsFirstModel}), Michael adds an additional independent variable, \group.},
        linerange={9},
        firstnumber=9,
        label={lst:michaelsSecondModel}
    ]{code/usage_scenario_group_exercise_statsmodels.py}
}

\newcommand{\groupExerciseCode}{
    \lstinputlisting[language=Python, caption={Example Tisane program from usage scenario (\autoref{sec:usage_scenario}).
    A Tisane program consists of a set of observed variables, expresses relationships between them, and queries Tisane for a statistical model by specifying a study design.
    Based on this input program, Tisane involves analysts in a disambiguation process (see~\ref{fig:groupExerciseDisambiguation}) to generate a final output statistical modeling script.}, label={lst:groupExerciseCode}]{code/usage_scenario_group_exercise.py}
}

\newcommand{\groupExerciseCodeVariables}{
    \lstinputlisting[language=Python,caption={The first snippet of the example Tisane program, written in the \SDSLlong, from the usage scenario (\autoref{sec:usage_scenario}). After importing \texttt{tisane}, Bridget specifies her observed variables.}, linerange={1-8},label={lst:groupExerciseCodeVariables}]{code/usage_scenario_group_exercise.py}
}

\newcommand{\groupExerciseCodeRelationships}{
    \lstinputlisting[language=Python,caption={A continuation of the snippet in~\autoref{lst:groupExerciseCodeVariables}. Bridget specifies the relationships between her observed variables using the methods \texttt{causes}, \texttt{associates\_with}, and \texttt{nests\_within.}}, linerange={10-12},firstnumber=10,label={lst:groupExerciseCodeRelationships}]{code/usage_scenario_group_exercise.py}
}

\newcommand{\groupExerciseCodeDesignAndQuery}{
    \lstinputlisting[language=Python,caption={The final snippet of the example Tisane program, continuing from~\autoref{lst:groupExerciseCodeRelationships}. Bridget queries Tisane for a statistical model by specifying her study design's most important variables and assigning data. Based on the complete program, consisting of this listing as well as~\autoref{lst:groupExerciseCodeVariables} and~\autoref{lst:groupExerciseCodeRelationships}, Bridget engages in a disambiguation process (see~\autoref{fig:groupExerciseDisambiguation}) to generate a final output statistical modeling script.}, linerange={13-15},firstnumber=13,label={lst:groupExerciseCodeDesignAndQuery}]{code/usage_scenario_group_exercise.py}
}

\newcommand{\groupExerciseOutputModel}{
    \lstinputlisting[
        language=Python,
        caption={Bridget's output statistical model from using Tisane. Tisane suggests a GLMM with group as a random intercept. The output script contains code for fitting this model and inspecting it using a residuals plot.},
        linerange={18-22},
        label={lst:groupExerciseOutputModel}
    ]{code/usage_scenario_tisane_output.py}
}

\newcommand{\groupExerciseDisambiguation}{
    \begin{figure}
        \centering
        \includegraphics[width=.95\linewidth]{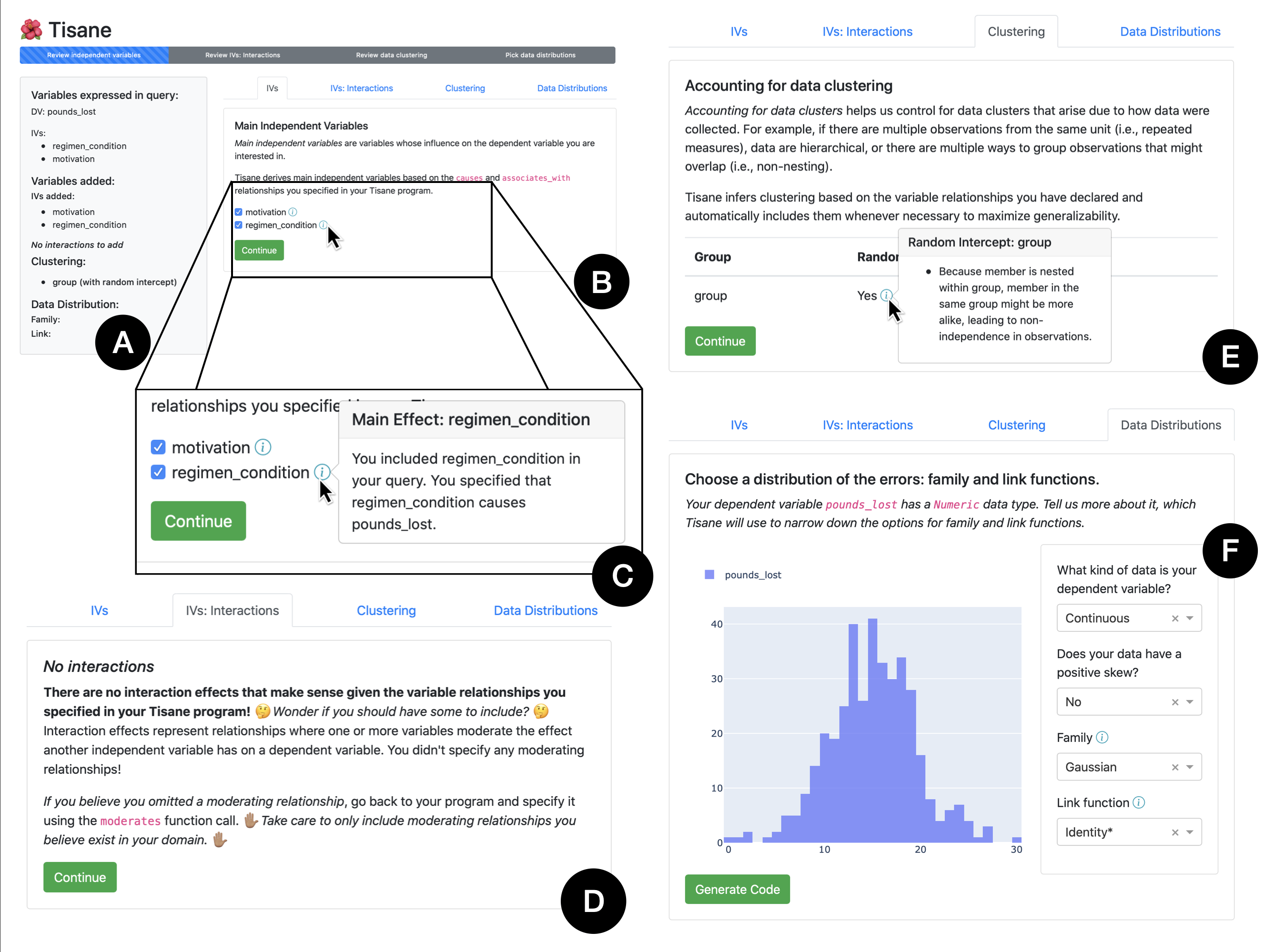}
        \caption{Example Tisane GUI for disambiguation from usage scenario. Tisane asks analysts disambiguating questions about variables that are conceptually relevant and that analysts may have overlooked in their query.
        (A) The left hand panel gives an overview of the model the analyst is constructing.
        (B) Based on the variable relationships analysts specify (\autoref{lst:groupExerciseCodeRelationships}), Tisane infers candidate main effects that may be potential confounders. Tisane asks analysts if they would like to include these variables, explaining in a tooltip
        (C) why the variable may be important to include.
        (D) Tisane only suggests interaction effects if analysts specify moderating relationships in their specification. This way, Tisane ensures that model structures are conceptually justifiable.
        (E) From the data measurement relationships analysts provide (line 15 in~\autoref{lst:groupExerciseCodeRelationships}), Tisane automatically infers and includes random effects to increase generalizability and external validity of statistical findings.
        (F) \diff{Tisane assists analysts in choosing an initial family and link function by asking them a series of questions about their dependent (e.g., Is the variable continuous or about count data?). To help analysts answer these questions and verify their assumptions about the data, Tisane shows a histogram of the dependent variable.}
        }
        \label{fig:groupExerciseDisambiguation}
        \Description{Five screenshots of the Tisane GUI are shown. The first screenshot has two labels, (A) and (B), which indicate the overview panel and the main independent variables panel respectively. Above the main effects panel is a series of tabs, which read from left to right: IVs, IVs: Interactions, Clustering, and Data Distributions. The second screenshot, which is labeled (C), shows the result of hovering over one of the info icons in the main independent variables panel, specifically the info icon next to regimen_condition. The tooltip’s body reads: “You included regimen_condition in your query. You specified that regimen_condition causes pounds_lost.” The third screenshot, labeled (D), shows the “IVs: Interactions” tab. It shows the following text, because there were no interaction effects to include in the model: “No interactions  There are no interaction effects that make sense given the variable relationships you specified in your Tisane program! 🤔Wonder if you should have some to include? 🤔 Interaction effects represent relationships where one or more variables moderate the effect another independent variable has on a dependent variable. You didn't specify any moderating relationships! If you believe you omitted a moderating relationship, go back to your program and specify it using the moderates function call. ✋🏽Take care to only include moderating relationships you believe exist in your domain. ✋🏽” The fourth screenshot, labeled (E), shows the “Clustering” tab. Text at the top of the tab’s panel reads: “Accounting for data clustering  Accounting for data clusters helps us control for data clusters that arise due to how data were collected. For example, if there are multiple observations from the same unit (i.e., repeated measures), data are hierarchical, or there are multiple ways to group observations that might overlap (i.e., non-nesting). Tisane infers clustering based on the variable relationships you have declared and automatically includes them whenever necessary to maximize generalizability.” Below the text is a table. There are two columns. The headers of the columns are “Group” and “Random Intercept.” There is one row beneath the header row, containing “group”, one of the variables in the usage scenario, and “Yes (info-icon)”. A mouse is depicted hovering over the (info-icon) next to “Yes”, and a tooltip has popped up. The header of the tooltip says “Random Intercept: group”. The body of the tooltip gives an explanation for why group was added as a random intercept: “Because member is nested within group, member in the same group might be more alike, leading to non-independence in observations.” Here, member and group are both data variables. The fifth screenshot shows the Data Distributions tab. At the top of the tab’s panel is the title text: “Choose a data distribution: family and link functions. Your dependent variable pounds_lost has a Numeric data type. Tell us more about it, which Tisane will use to narrow down the options for family and link functions.” A right hand panel (F) asks questions about the dependent variable (What kind of data is your dependent variable? Where the option Continuous is selected). To assist, a histogram of the dependent variable is shown. Based on the answers to these questions, Tisane suggests family and link functions. At the very bottom is a “Generate Code” button.}
      \end{figure}

}

\newcommand{\figurePossibleConfoundingAssociation}{
    \begin{figure}[H]
        \centering
        \includegraphics[width=.75\linewidth]{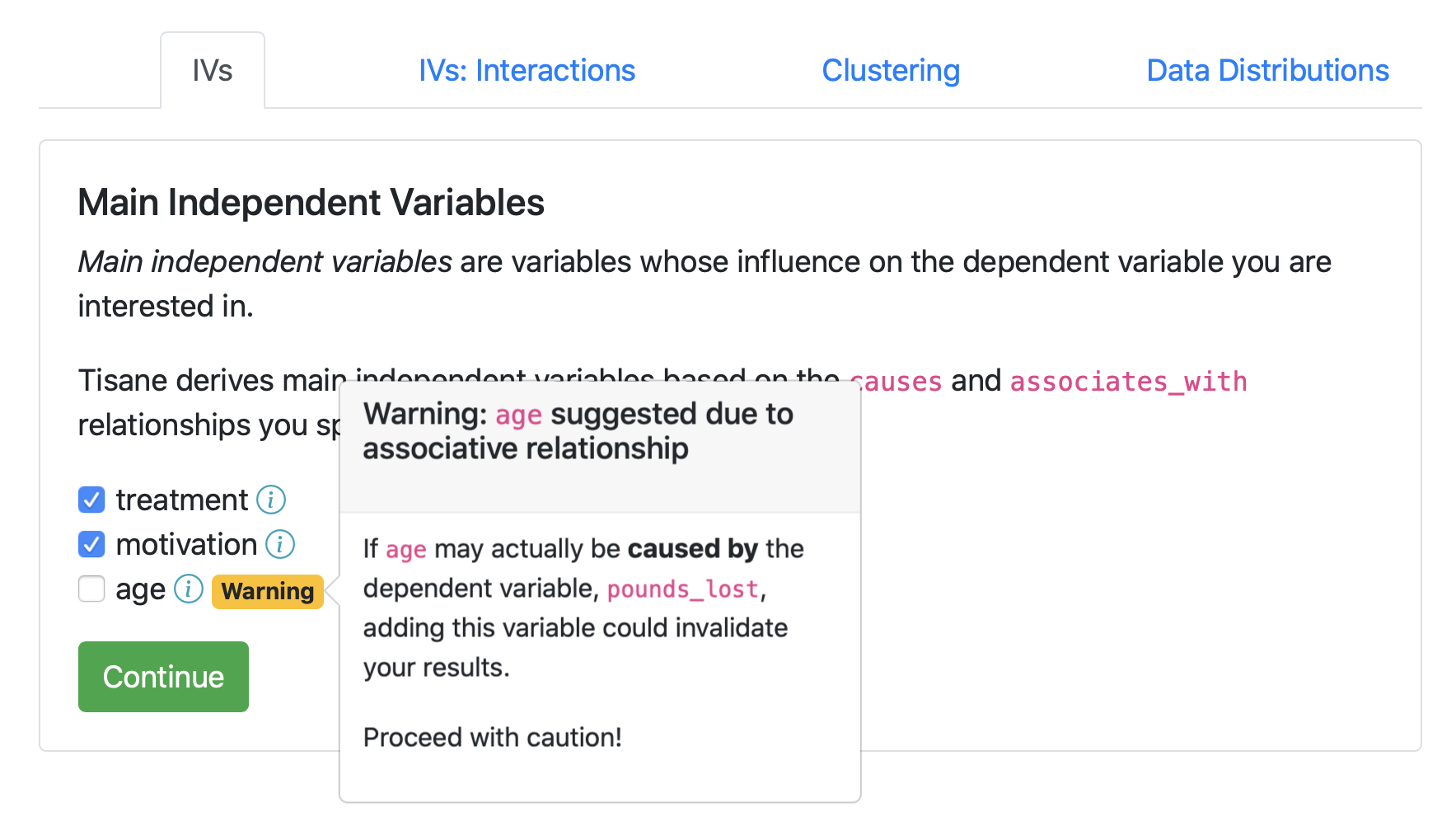}
        \caption{An example of the warning text given for potential confounding associations. When analysts hover over the ``Warning'' badge, a tooltip pops up that explains that they should be careful about adding this variable. Associative relationships may in actuality be causal relationships, and if in fact \texttt{pounds\_lost} \textbf{caused} \texttt{age}, then adding \texttt{age} would invalidate the model.}
        \label{fig:figurePossibleConfoundingAssociation}
    \end{figure}
}

\newcommand{\tableStudyDesignTools}{
  \begin{table}
    \small
    \caption{Overview of study design tools that informed Tisane's \SDSLlong.
    The first five tools provide higher-level
    abstractions. They are designed to help researchers reason about their study
    designs more holistically. The latter eight tools are lower-level and are more focused on stimuli,
    trials, and progressions between trials. *JsPsych is the base package to which JsPsychR, xprmtnr, and
    Jaysire provide wrappers and extensions.}
    \begin{tabular}{p{0.25\linewidth}|p{0.75\linewidth}}
      \textbf{Tool}	&	\textbf{Support provided} \\
      \hline
      Edibble~\cite{edibble} 	&	reason about end-to-end experimental design, create data collection schema  \\
      JMP Design of Experiments~\cite{jmpDOE}	&	use templates for experiments, some design optimization, some help with modeling \\
      Gosset~\cite{gosset}	&	search for optimal study design \\
      DeclareDesign~\cite{blair2019declareDesign} 	&	simulate data, specify and reason about designs statistically \\
      Touchstone2~\cite{eiselmayer2019touchstone2}	&	design controlled experiments while reasoning about randomization and statistical power	\\
      Formr~\cite{arslan2020formr}	&	design online survey questions and flow 	\\
      psychTestR~\cite{harrison2020psychtestr}	&	create trials, specify "timelines" for how trials should progress	\\
      Psychopy~\cite{peirce2007psychopy}	&	control how (visual) stimuli are presented, trials, and trial progression in an online experiment	\\
      Psychtoolbox~\cite{borgo2012psychtoolbox}	&	control stimuli in an online experiment, especially for neuroscience 	\\
      JsPysch*~\cite{de2015jspsych} & \multirow{4}{*}{create and control trials and stimuli for online experiments} \\
      JsPyschR~\cite{jspsychR} & \\
      xprmtnr~\cite{xprmntr} & \\
      Jaysire~\cite{jaysire} & \\
    \end{tabular}
    \label{tab:tableStudyDesignTools}
  \end{table}
}

\newcommand{\tableOverallCounts}{
  \begin{table}[ht]
    \begin{tabular}{|p{2cm}|c|c|c|c|c|c|} \hline
                        & \textbf{2017} &\textbf{2018} &\textbf{2019} &\textbf{2020} &\textbf{2021} &\textbf{Total} \\ \hline
      \textbf{repeated}                              &	130	&	168	&	173	&	171	&	163	&	805	\\ \hline
      \textbf{regression}                            &	49	&	66	&	71	&	72	&	72	&	330	\\ \hline
      \textbf{linear model}                          &	9	&	16	&	15	&	12	&	15	&	67	\\ \hline
      \textbf{generalized linear model\footnotemark} &	3	&	1	&	2	&	3	&	0	&	9	\\
      \hline
    \end{tabular}
    \caption{The number of papers from each year, and the total, that contained at least one instance of each of the key words.}
  \end{table}
  \footnotetext{(count is redundant in  ``linear model'')}
}

\newcommand{\tableLinearModelCounts}{
  \begin{table}[ht]
    \begin{tabular}{|l|c|} \hline
                  &	\multicolumn{1}{p{4cm}|}{Number of unique papers that use
                  ``regression'' and/or ``linear model''}	\\ \hline
    \textbf{2017}	&	52	\\ \hline
    \textbf{2018}	&	73	\\ \hline
    \textbf{2019}	&	78	\\ \hline
    \textbf{2020}	&	77	\\ \hline
    \textbf{2021}	&	81	\\ \hline
    \textbf{Total}  &   361 \\ \hline
    \end{tabular}
    \caption{The number of unique papers that contained the key word ``regression'' or ``linear model''.}
    \label{tab:reglmonly}
  \end{table}

}

\newcommand{\tableAnova}{
    \begin{table}
        \caption{Number of papers containing either of the key phrases ``ANOVA'' and ``analysis of variance''.}
        \begin{tabular}{ll} \hline
            \textbf{Year}   &	\textbf{Number of papers} \\ \hline
            2017            &   74 \\
            2018 &              87 \\
            2019 &              89 \\
            2020 &              108 \\
            2021 &              78 \\
            \textbf{Total} &    \textbf{436}
        \end{tabular}
        \label{tab:tableAnova}
    \end{table}
}

\newcommand{\michaelSecondModelOutput}{
    \lstinputlisting[
        language={},
        caption={Output for Michael's second model with pounds lost as the dependent variable and \regimen, \motivation, and \group as independent variables.},
        linerange={1-57},
        label={lst:michaelSecondModelOutput}
    ]{../output/no_interaction_effects.txt}
}

\newcommand{\bridgetModelOutput}{
    \lstinputlisting[
        language={},
        caption={Output for Bridget's model with pounds lost as the dependent variable, \regimen and \motivation as independent variables and \group as a random intercept.},
        linerange={60-81},
        label={lst:bridgetModelOutput}
    ]{../output/no_interaction_effects.txt}
}
\section{Introduction}



Statistical models play a critical role in how
people evaluate data and make decisions.
Policy makers rely on models to track disease, inform health
recommendations, and allocate resources. Scientists use models to
develop, evaluate, and compare theories. Journalists report on new findings in science,
which individuals use to make decisions that impact their nutrition, finances, and other aspects of their lives.
Faulty statistical models can lead to
spurious estimations of disease spread, findings that do not generalize or
reproduce, and a misinformed public. The challenge in developing accurate statistical models lies not in
a lack of access to mathematical tools, of which there are many
(e.g., R~\cite{team2013r}, Python~\cite{sanner1999python}, SPSS~\cite{spss}, and
SAS~\cite{sas}), but in accurately applying them in conjunction with
domain theory, data collection, and statistical
knowledge~\cite{jun2021hypothesisFormalization, mcelreath2020statistical}.

There is a mismatch between the interfaces existing statistical tools provide
and the needs of analysts, especially those who have domain knowledge but lack
deep statistical expertise (e.g., many researchers).
Current tools separate
reasoning about domain theory, study design, and statistical models, but analysts
need to reason about all three \emph{together} in order to author accurate
models~\cite{jun2021hypothesisFormalization}. 
For example, consider a researcher developing statistical models of hospital expenditure to
inform public policy. They collect data about individual hospitals within
counties. Based on their domain knowledge, they know that counties
have different demographics and that hospitals in these counties have
different funding sources (private vs. public), all of which influence hospital
spending. To model 
county-level and hospital-level attributes, the
researcher may author a generalized linear mixed-effects model (GLMM)
that accounts for clustering within counties. 
But which variables should they include?
How do they account for this
clustering? The three most common mistakes in modeling
hierarchical data~\cite{cohen2013applied} lead to miscalibrated statistical power, ``ecological
fallacies''~\cite{robinson1950}, and/or results that may not generalize, which
impact not only the validity of research findings~\cite{barr2013random} but also
enacted policies. How can the researcher avoid these issues?




To reduce threats to validity and improve analytical practices, \textbf{how
might we derive (initial) statistical models from knowledge about concepts and
data collection?} Inferring a statistical model raises two challenges: (1) How do we elicit
the information necessary for inferring a statistical
model? 
and (2) How do we infer a statistical model, given this information? We present
\textbf{Tisane, a system for integrating conceptual relationships, data
collection details, and modeling choices when specifying generalized linear
models (GLMs) and generalized linear mixed-effects models (GLMMs)}. GLMs and GLMMs are 
meaningful targets because they are commonly used (e.g., in
psychology~\cite{lo2015transform,cohen2013applied}, social
science~\cite{kreft1998introducing}, and
medicine~\cite{bolker2009generalized,barr2013random}) yet are easy to misspecify
for statistical experts and non-experts alike~\cite{barr2013random,
cohen2013applied}. We designed Tisane to support researchers who are domain
experts capable of supplying conceptual and data collection information but lack the statistical expertise or confidence to author GLM/GLMMs accurately.

Tisane provides a \textbf{\SDSLlong} for expressing 
relationships between variables. For example, the public health researcher can
express that average county income is associated with hospital spending based on
health economics theory or specify that hospitals exist within counties.
Tisane compiles the explicitly stated relationships into an internal
\textbf{graph representation} and then traverses the graph to infer candidate
GLMs/GLMMs. In this process, Tisane engages analysts in \textbf{interactive
compilation}. Analysts can query Tisane for a statistical model that
explains a specific dependent variable from a set of independent variables.
Based on the input query, Tisane asks analysts disambiguating questions to
output a script for fitting a valid GLM/GLMM.
Interactive compilation enables analysts to focus on their primary
variables of interest as the system checks that analysts do not overlook
relevant variables, such as potential confounders or data clustering that could
compromise generalizability. 
Figure~\ref{fig:figureSystemOverview} provides an overview
of this process.



To examine how Tisane affects real-world analyses, we conducted \textbf{case
studies with three researchers}. The researchers described how Tisane focused them on
their research goals, made them aware of domain assumptions, and helped
them avoid past mistakes. 
Tisane even helped one researcher correct their model prior to submitting to the ACM
Conference on Human Factors in Computing (CHI). 
These findings corroborate those
from an earlier pilot study that informed our design process (see supplemental material).

We contribute (1) a \SDSLlong and graph representation for recording and
reasoning about conceptual relationships between variables and data collection
procedures (\autoref{sec:dsl}), (2) an interactive compilation process that asks
disambiguating questions and outputs code for fitting and visualizing a GLM/GLMM
(\autoref{sec:interaction_model}), and (3) three case studies with researchers
that demonstrate the feasibility and benefit of prioritizing variable
relationships to author linear models (\autoref{sec:eval_with_users}). We also
provide an open-source Python implementation of Tisane.\footnote{Tisane is
available for download on pip, a popular Python package manager. The source code
is available at https://github.com/emjun/tisane.}


\figureSystemOverview

\section{Background and Related Work}
We first provide brief background about data analysis practices, GLMs/GLMMs, and causal
analysis. Then, we discuss how Tisane extends prior work on tools
for conceptual reasoning, study design, and automated statistical analysis.

\subsection{Data Analysis Practices}
Studies with analysts have found that data analysis is an iterative process that
involves data collection; cleaning and wrangling; and statistical testing and
modeling~\cite{grolemund2014cognitive,liu2019paths,liu2019understanding}. To
formalize their hypotheses as statistical model programs, analysts engage in a
dual-search process involving refinements to their conceptual understanding and
iterations on model implementations, under constraints of data and statistical
knowledge~\cite{jun2021hypothesisFormalization}. Analysts incorporate and refine
their 
domain knowledge, study
design, statistical models, and computational instantiations of statistical
models while creating statistical model programs. Tisane facilitates one formalization cycle in this
iterative process: deriving statistical models from conceptual knowledge and
data measurement specifications.


\subsection{Generalized Linear Models and Generalized Linear Mixed-effects Models} \label{sec:GLM}

Tisane supports two classes of models that are widely applicable to diverse
domains and data collection
settings~\cite{lo2015transform,barr2013random,bolker2009generalized}:
Generalized Linear Models (GLMs) and Generalized Linear Mixed-effects Models
(GLMMs). Both GLMs and GLMMs consist of (i) a \textit{model effects structure},
which can include main and interaction effects and (ii) \textit{family} and
\textit{link} functions. The family function describes how the residuals of a
model are distributed. The link function transforms the predicted values of the
dependent variable. This allows modeling of linear and non-linear relationships
between the dependent variable and the predictors. In contrast to
transformations applied directly to the dependent variable, a link function does
not affect the error distributions around the predicted values. The key
difference between GLMs and GLMMs is that GLMMs contain random effects in their
model effects structure. Random effects describe how individuals (e.g., a study
participant) vary and are necessary in the presence of hierarchies, repeated
measures, and non-nesting composition
(\autoref{sec:data-measurement-relationships})\footnote{Traditionally, the term
``mixed effects'' refers to the simultaneous presence of ``fixed'' and
``random'' effects in a single model. We try to avoid these terms as there are
many contradictory usages and definitions~\cite{gelmanFixedRandom}. When we do
use these terms, we use the definitions from Kreft and De
Leeuw~\cite{kreft1998introducing}.}.

Both GLMs and GLMMs assume that (i) the variables involved are linearly related,
(ii) there are no extreme outliers, and (iii) the family and link functions are
correctly specified. In addition, GLMs also assume that (iv) the observations
are independent. Tisane's interactive compilation process guides users through
specifying model effects structures, family and link functions to satisfy
assumption (iii), and random effects only when necessary to pick between GLMs
and GLMMs and satisfy assumption (iv).


\subsection{Causal Analysis}
There are multiple frameworks for reasoning about
causality~\cite{rubin2004teaching,pearl1995causal}. One widespread approach is
to use directed acyclic graphs (DAGs) to encode conditional dependencies between
variables~\cite{pearl1995doCalculus,greenland1999causal,spirtes1994conditional,spirtes1996using}.
If analysts can specify a formal causal graph, Pearl's ``backdoor path
criterion''~\cite{pearl1995causal,pearl2000models} explains the set of variables
that control for confounding. However, in practice, specifying proper causal
DAGs is challenging and error-prone for domain experts who are not also experts
in causal analysis~\cite{suzuki2020causal} due to uncertainty of empirical
findings~\cite{suzuki2018mechanisms} and lack of guidance on which variables and
relationships to include~\cite{velentgas2013developing}. Accordingly, Tisane
does not expect analysts to specify a formal causal graph. Instead, analysts can
express causal relationships as well as ``looser'' association (not causal)
relationships between variables in the \SDSLlong.


Prior work in the causal reasoning literature shows how linear models can be
derived from causal graphs to make statistical inferences and test the
motivating causal graph~\cite{spirtes1996using,spirtes1994conditional}.
Recently, VanderWeele proposed the ``modified disjunctive cause
criterion''~\cite{vanderweele2019modifiedDisjunctiveCriterion} as a new
heuristic for researchers without a clearly accepted formal causal model to
identify confounders to include in a linear model, for example. The criterion
identifies confounders in a graph based on expressed causal relationships.
Tisane applies the modified disjunctive cause criterion when suggesting
variables to include in a GLM or GLMM. Tisane does not automatically
include variables to the statistical models because substantive domain knowledge
is necessary to resolve issues of temporal dependence between variables, among
other considerations~\cite{vanderweele2019modifiedDisjunctiveCriterion}. To
guide analysts through the suggestions, Tisane provides analysts with explanations
to aid their decision making during disambiguation.

Finally, GLMs are not formal causal analyses. Tisane does not calculate average
causal effect or other causal estimands. Rather, Tisane only utilizes insights
about the connection between causal DAGs and linear models to guide analysts
towards including potentially relevant confounders in their GLMs grounded in domain knowledge. 



\subsection{Tools for Conceptual Reasoning and Study Design}

Tools such as Daggity~\cite{textor2011dagitty} support authoring, editing, and
formally analyzing causal graphs through code and a visual editor. Daggity
requires users to specify a formal causal graph, which may not always be
possible~\cite{suzuki2020causal,suzuki2018mechanisms,velentgas2013developing}.
Although a knowledgeable analyst could use Daggity to identify a set of
variables that control for confounding to include in a linear model, Daggity
does not provide this support directly. In contrast, Tisane aims to (i) help
analysts may not be able to formally specify causal graphs and (ii) scaffold the
derivation of GLMs and GLMMs from causal graphs.


Several domain-specific languages~\cite{gosset,bakshy2014planout} and tools
specialize in experiment
design~\cite{edibble,blair2019declaring,eiselmayer2019touchstone2}. A primary
focus is to provide researchers low-level control over trial-level and
randomization details. For example, JsPsych~\cite{deLeeuw2015jspsych} gives
researchers fine-grained control over the design and presentation of stimuli for
online experiments. At a mid-level of abstraction,
Touchstone~\cite{mackay2007touchstone} is a 
tool for designing and launching online experiments. It also refers users to R
and JMP for data analysis but does not help users author an appropriate
statistical model. Touchstone2~\cite{eiselmayer2019touchstone2} helps
researchers design experiments based on statistical power. At a high-level of
abstraction, edibble~\cite{edibble} helps researchers plan their data collection
schema. Edibble aims to provide a ``grammar of study design'' that focuses users
on their experimental manipulations in relation to specific units (e.g.,
participants, students, schools), the frequency and distribution of conditions (e.g.,
within-subjects vs. between-subjects), and measures to collect (e.g., age, grade,
location) in order to output a table to fill in during
data collection. While Tisane's \SDSLlong uses an abstraction level comparable
to edibble, Tisane is focused on using the expressed data measurement
relationships to infer a statistical model. Additionally, Tisane's \SDSL provides
conceptual relationships that are out of the scope of edibble but important for
specifying conceptually valid statistical models.








\subsection{Tools for Automated Statistical Analysis}


Researchers have introduced tools that automate statistical analyses.
Given a dataset, the Automatic Statistician~\cite{lloyd2014automatic} generates
a report listing all ``interesting'' relationships (e.g., correlations, statistical
models, etc.). 
Although apparently complete, the Automatic
Statistician may overlook analyses that are conceptually interesting and
difficult, if not impossible, to deduce from data alone.
In contrast, Tisane prioritizes analyst-specified conceptual and
data measurement relationships and uses them to bootstrap the modeling process.
As a result, Tisane aims to ensure that statistical analyses are not only
\emph{technically} correct but also \emph{conceptually} correct.

AutoML tools automate machine learning for non-experts. Tools such as
Auto-WEKA~\cite{autoweka}, auto-sklearn~\cite{autosklearn}, and H2O
AutoML~\cite{H2OAutoML20} aim to make statistical methods more widely usable.
Tisane differs from AutoML efforts in its focus on analysts who prioritize
explanation, not just prediction, such as researchers developing scientific
theories. As a result, Tisane provides support for specifying GLMMs, which some
prominent AutoML tools, such as auto-sklearn~\cite{autosklearn}, omit. Tisane
ensures that inferred statistical models respect expressed conceptual
relationships. Thus, Tisane programs can serve a secondary purpose of recording
and communicating conceptual and data measurement assumptions.
In addition,
Tisane explains its suggestions to users and guides them in answering
disambiguation questions whereas AutoML tools do not by default. Tisane's
explanations are grounded in the variable relationships analysts specify.
Although H2O AutoML offers a model explainability
module~\cite{H2OAutoMLexplainability}, these ``explanations'' take the form
of plots without conceptual exposition.

Tea~\cite{jun2019tea} elicits
end-user expertise through explicit hypotheses and study designs to
automatically infer a set of valid Null Hypothesis Significance Tests. Tisane
differs from Tea in three key ways. First, Tisane enables analysts to express
more complex study designs, such as nested hierarchies that necessitate mixed
effects modeling, which are 
not available in Tea. Second,
because statistical modeling requires more conceptual expertise and oversight,
Tisane is a mixed-initiative system while Tea is not. Third,
Tisane outputs a single statistical model, whereas Tea outputs a set of
statistical tests.


Recent work in the database community helps researchers answer causal questions
about multilevel, or hierarchical, data~\cite{salimi2020causal, kayali2020demonstration}.
CaRL~\cite{salimi2020causal} provides a domain-specific language to express
causal relationships between variables and a GUI to show researchers 
results.
Like CaRL, Tisane leverages the insight that
researchers have domain knowledge that a system can use 
to
infer statistical methods.
Whereas CaRL is focused on answering specific queries about average causal
effect, Tisane supports authoring GLMs and GLMMs, which can address a range of non-causal questions.

In summary, whereas prior systems have supported reasoning about either
concepts, study designs, or statistical models, Tisane integrates all three.
Existing theories of data analysis~\cite{jun2021hypothesisFormalization,
mcelreath2020statistical} illustrate how all three concerns are necessary and
interconnected. Thus, a tool for integrating these concerns seems a promising
way to reduce errors and the cognitive burden involved in model
specification.


\def\poundslost{\texttt{pounds\_lost}\xspace}
\def\motivation{\texttt{motivation}\xspace}
\def\regimen{\texttt{regimen\_condition}\xspace}
\def\regimencondition{\texttt{regimen\_condition}\xspace}
\def\group{\texttt{group}\xspace}
\def\age{\texttt{age}\xspace}
\section{Usage Scenario} \label{sec:usage_scenario}


To illustrate how a researcher might use Tisane, we compare two hypothetical
researchers analyzing the same dataset. This scenario (simplified
from~\cite{cohen2013applied}) illustrates workflow differences between Tisane
and current tools (\statsmodels in Python and \texttt{lmer} in R). Michael and
Bridget are health experts studying the effects of a new exercise regimen they
have developed on weight loss. Their research question is ``How much does the exercise regimen affect weight loss?'' They recruited 386 adults to be part of 40
exercise groups focused on diet and weight management. The researchers randomly
assigned 16 groups a control regimen and the other 24 groups an experimental
regimen. The researchers measured the adults' motivation scores for weight loss
at the beginning of the experiment and their total weight loss at the end of the
experiment. Michael uses \statsmodels~\cite{statsmodelsRef} to analyze the data\footnote{The workflow in R is almost
identical to the workflow in Python using
\statsmodels.}. Bridget uses Tisane.
While both are experienced researchers, familiar with their shared field of
study, Michael and Bridget are not statistics experts. They have both used GLMs
in the past but neither has heard of GLMMs.

\subsection{Workflow in Python using \statsmodels}
Michael takes a first attempt at creating a model. He loads the data and
casts \regimen and \group as categorical variables. The first model
(\autoref{lst:michaelsFirstModel}) that Michael tries has the dependent variable
\poundslost and the independent variables \regimen (control vs.
treatment) and \motivation.

\michaelsFirstModel


Although this model includes the primary variable of research
interest (\regimen) and a likely confounder (\motivation),
Michael realizes that this model overlooks the fact that the adults exercised in
groups. The groups likely had group support and accountability, among other
benefits of group cohesion that are difficult to measure and not included in the
data.


In his second model (\autoref{lst:michaelsSecondModel}), Michael
adds \group as an additional independent variable. Considering that this
model accounts for variables pertaining to individual adults (i.e., \poundslost, \motivation) and to groups (i.e., \regimen), Michael supposes this model is good enough and accepts
it as his final model.

\michaelsSecondModel


\subsection{Workflow with Tisane}
\renewcommand\regimen{\texttt{regimen\_condition}\xspace}
\def\adult{\texttt{adult}\xspace}
Bridget starts by listing the variables of interest. Lines 3-8 in
Listing~\ref{lst:groupExerciseCodeVariables} show how variables are declared
with a data type, the name of the column in the data that corresponds to the
variable.
The observational units are \adult and \group. The measures
\motivation and \poundslost pertain to individual \adult{}s while
exercise \regimen was administered to exercise \group{}s. The Tisane program
makes these details explicit. Because Bridget has data, she does not need to
specify the cardinality of variables. \groupExerciseCodeVariables

By expressing how the variables relate to one another, Bridget becomes more
consciously aware of the assumptions the research team has made about their
domain. In line 10 of~\autoref{lst:groupExerciseCodeRelationships}, Bridget specifies that \regimen directly causes
\poundslost while in line 11 \motivation is associated with, but
not necessarily a cause of, \poundslost.
She also expresses that
\adult{}s were nested in \group{}s in line 12.
\groupExerciseCodeRelationships

Next, in line 14 of~\autoref{lst:groupExerciseCodeDesignAndQuery}, Bridget specifies a study design with \poundslost as the dependent variable
and \regimen and \motivation as the independent variables of interest and assigns data to the design. 
Bridget uses this design to query Tisane for a statistical model in line 15.
\groupExerciseCodeDesignAndQuery

\groupExerciseDisambiguation

Tisane launches a GUI in the browser after executing the program.\footnote{As
explained in~\autoref{sec:disambiguation}, if Bridget had executed the script in
a Jupyter notebook, the GUI would open in the notebook.} In the GUI, Tisane
asks Bridget to look over her choice of variables
(\autoref{fig:groupExerciseDisambiguation}). As seen in panel B,
there are no additional variables to consider.
Bridget continues to the next tab, interaction effects
(panel D). Tisane explains that interaction effects do not make sense for her model given that she
did not express any moderating relationships. In this way, Tisane prevents her
from adding arbitrary interaction effects without conceptual foundation. 
In the next tab (E), Bridget
sees that Tisane has automatically included exercise \group as a random intercept. (Tisane does not include a random slope for group
because there is only one observation per adult in a group, see~\autoref{sec:deriveRandomEffects}.) 
Bridget has not heard of a random intercept before, but she reads the
explanation, which explains that accounting for exercise \group{}s is necessary
since \adult{}s were in \group{}s and \group{}s received the \regimen{}s. In the
last tab, Bridget answers questions about the the dependent variable \poundslost
to identify family and link functions. She specifies that the dependent variable is continuous. She chooses the Gaussian family with the default link function.
Finally, Bridget clicks on the button to generate code. Tisane's output script
contains code to fit the statistical model and plot model residuals to inspect
it. Tisane helped Bridget
author the following GLMM: \groupExerciseOutputModel

\subsection{Key differences in workflows and statistical results}

Even with experience modeling in Python, Michael makes
two common mistakes in authoring linear
models~\cite{cohen2013applied,kreft1998introducing}: disaggregating observations
and artificially inflating statistical power (model 1) and using a ``fixed
effects approach to clustering'' that compromises the generalizability of
findings~\cite{barr2013random} (model 2). Despite knowing all the pertinent
information about the domain and data collection for constructing a valid
statistical model, Michael is not able to leverage it without additional
statistical expertise, and, as a result, arrives at different statistical results than Bridget.

In contrast, Bridget and other analysts using Tisane express their knowledge using the 
\SDSLlong, answer a few disambiguating questions with guidance, and receive as output a script for executing a statistical model. Tisane
helps researchers like Bridget leverage their conceptual and data collection
knowledge to author statistical models that optimize for statistical conclusion
and external validity, avoiding errors and reducing the cognitive burden along the way.

\diff{Once fit, Michael's and Bridget's final models (see
\autoref{tab:usageScenarioModel}) disagree on the precise effect sizes of
\regimen and \motivation, which are pertinent to their motivating research question. Michael
concludes that \regimen and \motivation are less important than they really are. Additionally, the coefficients
and standard errors for each \group suggest that Michael's model overlooks
important group differences. Therefore, using a GLM instead of the appropriate model, a
GLMM, leads Michael to answer the research question differently than Bridget, artificially inflates statistical power~\cite{cohen2013applied}, and compromises the generalizability of his findings~\cite{barr2013random}.}\footnote{Another
common mistake, not shown here, is to aggregate observations and use group means
of the independent variables in the model, artificially deflating statistical
power (``ecological fallacy''~\cite{robinson1950}). Kreft and De
Leeuw~\cite{kreft1998introducing} share an example where disaggregating vs.
aggregating data lead to different signs for a fitted parameter. Unfortunately,
we could not access the data to illustrate this here.}

\begin{table}[h]
  \renewcommand{\arraystretch}{1.1}
  \begin{tabular}{|c|c|c|c|c|c|c|} \hline
            & $\beta_\regimencondition$  & $p$   & $\beta_\motivation$ & $p$  & $\beta_\group$ & $p$ \\ \hline
    Michael & 1.628            & .046  & 3.119              & .000 & Various        & Various \\ \hline
    Bridget & 1.659            & .005  & 3.193              & .000 & N/A            & N/A  \\ \hline
  \end{tabular}
  \caption{The coefficients for each of the independent variables in Michael's and Bridget's models. \diff{The complete output tables for
Michael's and Bridget's models are included in supplemental material.}}
  \label{tab:usageScenarioModel}
\end{table}

\section{Design goals} \label{sec:design_considerations}

We articulated four design goals based on prior research and our formative
work. The supplemental material details our design process.

	\dcConceptualKnowledgeLong Current tools require analysts to transition back
	and forth between their conceptual concerns and their statistical model
	specifications using math and/or code~\cite{jun2021hypothesisFormalization}.
	Analysts' conceptual knowledge remains implicit and
	hidden~\cite{mcelreath2020statistical}. As a result, analysts may resort to
	familiar but sub-optimal statistical
	methods~\cite{jun2021hypothesisFormalization} or accidentally overlook
	details that lead to conceptually inaccurate statistical models. One
	solution is to provide tools at a higher level of abstraction that allow
	analysts to express their conceptual knowledge directly. However, a higher
	level of abstraction alone is not enough. Tools must then leverage the
	expressed conceptual knowledge to guide analysis authoring.

	Tisane provides a high-level \SDSLlong 
	that captures the
	motivation behind a study (\autoref{sec:dsl}).
	Tisane represents the specification in an internal graph
	representation to derive only conceptually accurate statistical model
	candidates. To arrive at an output statistical model, Tisane asks analysts
	disambiguating questions and provides them with suggestions and explanations
	based on their expressed variable relationships
	(\autoref{sec:interaction_model}). Importantly, Tisane does not fit or show
	modeling results during disambiguation to discourage statistical fishing.
	Although Tisane does not prevent researchers from re-starting and iterating on their Tisane
	program to attain specific statistical model findings, Tisane programs act
	as documentation for 
	conceptual relationships that 
	others could audit.

	\dcValidityLong At present, the burden of valid statistics lies entirely on
	analysts. Tisane divides some of this burden by (i) ensuring correct
	application of methods (i.e., GLM vs. GLMM) and (ii) inferring models that
	increase the generalizability of results for
	GLMMs~\cite{barr2013random,barr2013randomUpdated}. Tisane helps analysts
	author GLMs and GLMMs that satisfy two assumptions: (i) observational
	dependencies and (ii) correct family and link functions. First, Tisane
	infers and constructs maximal random effects that account for dependencies
	due to repeated measures, hierarchical data, and non-nesting compositions.
	\diff{Maximal effects structures account for within-sample variability and
	thereby mitigate threats to external validity due to sampling biases from
	the choice of observational units and settings~\cite{shadish2010campbell}.}
	Second, Tisane narrows the set of viable family and link functions to match
	the dependent variable's data type (e.g., numeric). Tisane's GUI asks
	follow-up questions to determine the \textit{semantic} type of variables
	(e.g., counts), further narrowing analysts' family and link function
	choices. The output script also plots model residuals against fitted values
	and provides tips (as comments) for interpreting the plot. The family and
	link functions Tisane suggests are intended to bootstrap an initial
	statistical model that analysts can examine and, if necessary, revise. This
	is how Tisane helps analysts avoid four common threats to statistical
	conclusion and external validity~\cite{cook2002generalizedCausalInference}:
	(i) violation of statistical method assumptions, (ii) fishing for
	statistical results, (iii) not accounting for the influence of specific
	units, and (iv) overlooking the influence of data collection procedures on
	outcomes.
	\dcGuidanceLong Analysts may have insight into their research questions and
	domain that a system cannot capture. At the same time, analysts, especially those with
	less statistical experience, may lack the knowledge to select among many
	possible statistical models, which may inadvertently encourage
	cherry-picking based on observed results. Thus, Tisane adopts an interaction
	model that asks analysts specific questions to resolve modeling ambiguity
	rather than show multiple statistical models at the same time. Tisane also
	does not automatically select a a ``best'' model (e.g., highest R$^2$,
	easiest to interpret) but rather gives analysts suggestions and explanations
	to help them come to a statistical model that is valid and appropriate for
	their goals.

\diff{\dcStatisticalPlanningLong Experimental design best practices, such as
pre-registration, encourage researchers to plan their statistical analyses prior
to data collection. In accordance, Tisane does not require analysts to provide
data. If analysts do not have data, analysts must specify the cardinality of
variables at declaration. Without data, Tisane cannot validate variable
declarations but otherwise guide analysts through the same interactive
compilation process. The output Tisane script will include an empty file path
and a comment directing analysts to specify a data path prior to execution. 
Analysts could attach this output Tisane script to their pre-registrations. Once
analysts collect data, they can re-run their previously specified Tisane program
to validate and inspect their data. If Tisane does not issue any validation errors, analysts can proceed to generate an
identical script or provide a file path in the pre-registered script and execute
it.}

\section{\SDSLlong and graph representation} \label{sec:dsl}

Tisane provides a \textit{\SDSLlong} (\textit{\SDSL}) for expressing
relationships between variables. There are two key challenges in designing a
specification from which to infer statistical models: (1) determining the set of
relationships that are essential for statistical modeling and (2) determining
the level of granularity to express relationships.

In Tisane's SDSL, analysts can express conceptual and data measurement
relationships between variables. Both are necessary to specify the domain
knowledge and study designs from which Tisane infers statistical models.

\subsection{Variables}
There are three types of data variables in Tisane's \SDSL: (i) units, (ii)
measures, and (iii) study environment settings. The \textbf{Unit} type
represents entities that are observed and/or receive experimental treatments. In
the experimental design literature, these entities are referred to as
``observational units'' and ``experimental units,'' respectively. Entities can
be both observational and experimental units simultaneously, so the \SDSL does
not provide more granular unit sub-types. The \textbf{Measure} type represents
attributes of units and must be constructed through their units, e.g.,
\texttt{age = adult.numeric(\upquote{age})}. Measures are proxies (e.g., minutes
ran on a treadmill) of underlying constructs (e.g., endurance). Measures can
have one of the following data types: numeric, nominal, or ordinal. Numeric
measures have values that lie on an interval or ratio scale (e.g., age, minutes
ran on a treadmill). Nominal measures are categorical variables without an
ordering (e.g., race). Ordinal measures are ordered categorical variables (e.g.,
grade level in school). \diff{We included these data types because they are
commonly taught and used in data analysis.}
The \textbf{SetUp}
type represents study environment settings that are neither units nor measures. For
example, time is often an environmental variable that differentiates repeated
measures but is neither a unit nor a measure 
of a specific unit.



\subsection{Relationships between Variables}
In Tisane's \SDSL, variables have relationships that fall into two broad
categories: (1) \textit{conceptual relationships} that describe how variables
relate theoretically and (2) \textit{data measurement relationships} that
describe how the data was, or will be, collected. Below, we define each of the
relationships in Tisane' \SDSL and describe how Tisane
internally represents these relationships as a graph (as illustrated
in~\autoref{fig:figureSDSLToGraphIR}).~\autoref{fig:figureGraphIRExample} shows
the graph representation constructed from the usage scenario.

Tisane's graph IR is a 
directed multigraph.
Nodes
represent variables, and directed edges represent relationships between
variables. Tisane internally uses a graph intermediate representation (IR) because graphs are widely
used for both conceptual modeling and statistical analysis, two sets of
considerations that Tisane unifies.

Tisane's graph IR differs from two types of graphs
used in data analysis: causal DAGs and path analysis diagrams. Unlike
causal DAGs, Tisane's graph IR allows for non-causal relationships, moderating relationships
(i.e., interaction effects), and data measurement relationships that
are necessary for inferring random effects. Unlike path analysis diagrams that
allow edges to point to other edges to represent
interaction effects,
Tisane represents interactions as separate nodes and only allows nodes as endpoints
for edges. These design decisions simplify our statistical model
inference algorithms and their implementation.

\subsubsection{Conceptual relationships.}

Tisane's \SDSL supports three conceptual relationships: causes, associates with,
and moderates. Analysts can express that a variable \textbf{causes} or is
\textbf{associated with} (but not directly causally related to) another variable.
Variables associated with the dependent variable, for example, may help explain
the dependent variable even if the causal mechanism is unknown. \diff{If analysts are
aware of or suspect a causal relationship, they should use
\texttt{causes}.}

We chose to support both causal and associative relationships because formal
causal DAGs are difficult for domain experts to
specify~\cite{suzuki2020causal,suzuki2018mechanisms,velentgas2013developing},
prior work has observed that researchers already use informal graphs that
contain associative relationships when reasoning about their hypotheses and
analyses~\cite{jun2021hypothesisFormalization}, and GLMs/GLMMs can represent
non-causal relationships. Finally, analysts can also express interactions where
one (or more) variable (the \textit{moderating variables}) \textbf{moderates}
the effect of a \textit{moderated variable} on another variable (the
\textit{target variable}).



Mediation relationships (where one variable influences another through a middle variable) are another common conceptual relationship. Tisane does
not provide a separate language construct for 
mediation because mediations are expressible using two or more causal
relationships. Furthermore, mediation analyses require specific analyses, such
as structural equation modeling~\cite{hoyle1995SEM}, that are out of Tisane's
scope.


In the graph IR, a \texttt{causes} relationship introduces a causal edge from
one node, the cause, to another node, the effect (\autoref{fig:figureSDSLToGraphIR}(a)). Because a
variable cannot be both the cause and effect of the same variable, any pair of
nodes can only have one causal edge between them. Furthermore, from a formal
causal analysis perspective, associations may indicate the presence of a hidden,
unobserved variable that mediates the causal effect of a variable on another or
that influences two or more variables simultaneously. Thus, rather than
inferring or requiring analysts to specify hidden variables, which may be
unknown and/or unmeasurable, the \texttt{associates\_with} relationship introduces two directed edges in
opposing directions, representing the bidirectionality of association (\autoref{fig:figureSDSLToGraphIR}(b)). A \texttt{moderates}
relationship creates a new node that is eventually transformed into an interaction term in the model, introduces associative edges between the new
interaction node and the target (variable) node, creates associative edges between the moderated variable's node and the target node, and adds associative
edges between the moderating variables' nodes and the
target node if there is not a causal or associative edge already (\autoref{fig:figureSDSLToGraphIR}(c)).
Furthermore, each interaction node inherits the attribution edges from the nodes of the
moderating variables that comprise it. This means that every interaction node is
also the attribute of
at least one unit.\footnote{In statistical terms, this
means that within-level interactions have one unit while cross-level
interactions may have two or more units.}

\figureSDSLToGraphIR

\subsubsection{Data measurement relationships.}\label{sec:data-measurement-relationships}


Study designs may have clusters of observations that need to be modeled explicitly for external validity.
For example, in a within-subjects experiment, participants provide multiple
observations for different conditions. An individual's observations may cluster
together due to a hidden latent variable. Such clustering may be imperceptible
during exploratory data visualization of a sample but can threaten external validity.
GLMMs can mitigate three common sources of clustering that
arise during data collection
~\cite{gelmanHill2006regression,kreft1998introducing,cohen1988statistical}:

\begin{itemize}
  \item \textbf{Hierarchies} arise when one observational/experimental unit
  (e.g., adult) nests within another observational/experimental unit (e.g.,
  group). This means that each instance of the nested unit belongs to one and
  only one nesting unit (many-to-one).
  \item \textbf{Repeated measures} introduce clustering of observations from the
  same unit instance (e.g., participant).
  \item \textbf{Non-nesting composition} arises when overlapping attributes
  (e.g., stimuli, condition) describe the same observational/experimental unit
  (e.g., participant)~\cite{gelmanHill2006regression}.
\end{itemize}

The above sources of clustering pose three problems for analysts. First,
analysts must have significant statistical expertise to identify when data
observations cluster. Second, they must know how to mitigate these clusters in
their models. Third, with this knowledge, analysts must figure out how to
express these types of clustering in their analytical tools. Even if analysts
are not able to identify clustered observations, they are knowledgeable about
how data were collected.


Thus, Tisane addresses the three problems by (i) eliciting data measurement
relationships from analysts to infer clusters and (ii) formulating the maximal
random effects structure, optimizing for external validity
(\autoref{sec:interaction_model}). Below, we describe language features for expressing data measurement relationships.




\paragraph{Nesting relationships: Hierarchies}
\textbf{Hierarchies} arise when a unit (e.g., an \adult) is nested within another
unit (e.g., an exercise \group). Researchers may collect data with
hierarchies to study individual and group dynamics together or as a side effect of
recruitment strategies. To express such designs, Tisane provides the
\texttt{nests\_within} construct. Conceptually, nesting is strictly between
observational/experimental units, so Tisane type checks that the variables
that nest are both Units. 
In the graph IR, a nesting relationship is encoded as an edge between two unit
nodes (\autoref{fig:figureSDSLToGraphIR}(d)). There is one edge from the nested
unit (e.g., \adult) to the nesting unit (e.g., \group)~\footnote{\diff{The code base contains examples with nesting relationships.}}.

\paragraph{Frequency of measures: Repeated measures, Non-nesting composition}
\def\numberofinstances{\texttt{number\_of\_instances}\xspace}
When a measure is declared through a unit, Tisane adds an
attribution edge (``has') from a unit node to a measure node (\autoref{fig:figureSDSLToGraphIR}(e)).
A unit's measure can be taken one or more times in a study. The frequency of
measurement is useful for detecting repeated measures and non-nesting
composition. In \textbf{repeated measures} study designs, each unit provides
multiple values of a measure, which are distinguished by another variable,
usually time. \textbf{Non-nesting}~\cite{gelmanHill2006regression} composition
arises when measures describing the same unit overlap. For example, HCI researchers studying input devices might
design them to utilize different senses (e.g., touch, sight, sound).
Participants in the study may be exposed to multiple different devices, which
act as experimental conditions of senses. The conditions are intrinsically tied to the
devices, and participants can be described as having both conditions and
devices, which overlap with one another. Such study designs
introduce dependencies between observations~\cite{clark1973language} and hence
violate the assumption of independence that GLMs make.

\def\inputdevice{\texttt{device}\xspace} When analysts declare Measures, they
specify the frequency of the observation through the
\texttt{number\_of\_instances} parameter. This parameter accepts an integer,
variable, a Tisane \texttt{Exactly} operator, or a Tisane \texttt{AtMost}
operator. By default, the parameter is set to one. The \texttt{Exactly} operator
represents the exact number of times a unit has a measure. The \texttt{AtMost}
operator represents the maximum number of times a unit has a measure. Both
operators are useful for specifying that a measure's frequency depends on
another variable, which is expressible through the \texttt{per} function. For
example, participants may use two \inputdevice{}s \textit{per}
\texttt{condition} assigned: \texttt{device = subject.nominal(\upquote{Input
device}, number\_of\_instances=ts.Exactly(2).per(condition))}. The \texttt{per}
function uses the Tisane variable's cardinality by default but can instead use a
data variable's \numberofinstances by specifying \texttt{use\_cardinality=False}
as a parameter to \texttt{per}. Moreover, specifying a measure's
\texttt{number\_of\_instances} to be an integer is syntactic sugar for using the
\texttt{Exactly} operator. Specifying a variable is syntactic sugar for
expressing \texttt{ts.Exactly(1).per(variable)}.

To determine the presence of repeated measures or non-nesting composition,
Tisane computes the \numberofinstances of measures and their relationship to
other measures. Measures that are declared with \numberofinstances equal to one
are considered to vary between-unit. Measures that are declared with
\numberofinstances greater than one or a variable with cardinality greater than
one are considered to vary within-unit as repeated measures. If there are
instances of a measure per another measure sharing the same unit, the measures
are non-nesting.

\figureGraphIRExample

\section{Statistical model inference: Interactively querying the graph IR}
\label{sec:interaction_model} After specifying variable relationships, analysts
can query Tisane for a statistical model. Queries are constructed by specifying
a study design with a dependent variable (the value to be predicted) and a set
of independent variables (predictors). Tisane processes the query and generates
a statistical model in four phases: (1) preliminary conceptual checks that
validate the study design, (2) inference of possible effects structures and
family and link functions, (3) input elicitation to disambiguate possible
models, and (4) generation of a \diff{record of decisions during disambiguation
and the} final executable script. Given that the interactive process begins with
an input program using Tisane and outputs a script for fitting a GLM or GLMM, we
call this process \textit{interactive compilation}.


\subsection{Preliminary checks} \label{sec:prelim-checks}
At the beginning of processing a query, Tisane checks that every input study
design is well-formed. This involves two conceptual correctness checks. First,
every independent variable (IV) in the study design must either cause or be
associated with the dependent variable (DV) directly or transitively. Second, the
DV must not cause any of the
IVs, since it would be conceptually invalid to explain a
cause from any of its effects. If any of the above checks fail, Tisane
issues a warning and halts execution. By using these two checks, the Tisane
compiler avoids technically correct statistical models that have little to no
conceptual grounding (\dcConceptualKnowledge). If the checks pass, Tisane proceeds to the next phase.

\subsection{Candidate statistical model generation}
A GLM/GLMM is comprised of a model effects structure, family function, and link
function. The model effects structure may consist of main, interaction, and
random effects. Tisane utilizes variables' conceptual relationships to infer candidate
main and interaction effects and data measurement relationships to infer
random effects. Tisane infers family and link functions based on the data type
of the DV in the query. The candidate statistical models that Tisane
generates based on the graph and query seeds an interactive disambiguation
process.

The purpose of identifying candidate main effects beyond the ones analysts may
have specified is to provoke consideration of erroneously omitted variables that
are conceptually relevant and pre-empt potential confounding and
multicollinearity issues that may arise.

\subsubsection{Deriving Candidate Main Effects}
In a query to infer a statistical model, analysts specify a single dependent
variable and a set of one or more IVs. After passing the checks described in~\autoref{sec:prelim-checks}, 
the query's independent variables are considered candidates. In addition, Tisane
derives three additional sets of candidate main effects intended to control for
confounding variables in the output statistical model\footnote{Tisane currently
treats each input IV as a separate ``exposure'' variable for which to identify
confounders. Tisane then combines all confounders into one statistical model.}.
The first two sets below are from the ``modified disjunctive cause
criterion''~\cite{vanderweele2019modifiedDisjunctiveCriterion}:

\begin{itemize}

	\item \textbf{Causal parents.} For each IV in the query, Tisane finds its
	causal parents (see~\autoref{fig:figureCandidateMainEffects}(a)).
	\item \textbf{Possible causal omissions.} Tisane looks to see if any other
	variables not included as IVs cause the DV 
	(see
	in~\autoref{fig:figureCandidateMainEffects}(b)). They are relevant to the DV
	but may have been erroneously omitted.
	\item \textbf{Possible confounding associations.} For each IV, Tisane looks
	for variables that are associated with both the IV and the DV (see
	in~\autoref{fig:figureCandidateMainEffects}(c)). Because associations
	between variables can have multiple underlying causal structures, Tisane
	recommends variables with associative relationships with caution. Tisane
	issues a warning describing when not to include such a variable in the GUI (see Figure 3 in supplemental material).
\end{itemize}

\figureCandidateMainEffects

Using the above rules, Tisane suggests a set of variables that are likely
confounders of the variables of interest expressed in the query. There may be
additional confounders due to unmeasured or unexpressed variables that are either
not known or excluded from the graph. Tisane never automatically includes the
candidate main effects in the output statistical model. Analysts must always
specify a variable as an IV in the query or accept a suggestion (\dcGuidance).

If a graph only contains associates edges then the candidate main effects Tisane
suggests are those that are directly associated with both the DV and an IV. If a
graph has only causal edges, Tisane would suggest variables that directly cause
the DV but were omitted from the query and the causal parents of IVs in case the
parents exert causal influence on the DV through the IV or another variable that is not
specified.


The total set of main effects, including variables the analyst has specified as
IVs in their query and candidate main effects, are used to derive candidate interaction
effects and random effects, which we discuss next.

\subsubsection{Deriving Candidate Interaction Effects}
An interaction between variables means that the effect of one variable (the \textit{moderated} variable) on a \textit{target} variable is moderated by another (non-empty) set of variables (the \textit{moderating} variables). Tisane's
\SDSL already provides a primitive, \texttt{moderates}, to
express interactions. As such, Tisane's goal in suggesting candidate interaction
effects is to help analysts avoid omissions of conceptual relationships
that are pertinent to an analyst's research questions or hypotheses (\dcConceptualKnowledge).
Candidate interaction effects are the interaction nodes whose (i) moderated and moderating variables include two or more candidate main effects and (ii)
 target variable is the query's DV.


\subsubsection{Deriving Candidate Random Effects} \label{sec:deriveRandomEffects}
Random effects occur when there are clusters in the data, which occur when
we have repeated measures,
nested hierarchies, or non-nesting composition (as defined in Section~\ref{sec:data-measurement-relationships}). Tisane implements Barr et al.'s recommendations
for specifying the maximal random effects structure of linear mixed effects
models for increasing the generalizability of statistical
results~\cite{barr2013random, barr2013randomUpdated}.

To derive random effects, Tisane focuses on the data measurement edges in the
graph IR. Using the graph IR, Tisane identifies unit nodes, looks for any nesting
edges among them, and determines within- or between-subjects
measures based on the frequency of observations for units. 
From these, Tisane
generates random intercepts of units for the unit's measures that are between-subjects as well as the unit's measures that are within-subjects where each instance of the unit
has only one observation per value of another variable. Tisane generates random slopes of a unit and its measure for all measures
that are within-subjects where each instance of the unit has multiple
observations per value of another variable. For interaction effects, random
slopes are included for the largest subset of within-subjects variables (see~\cite{barr2013randomUpdated}). 
Tisane handles correlation of random slopes and intercepts during disambiguation (section~\ref{sec:disambiguation}).
Maximal random effects may lead to model convergence issues that analysts
address by later removing or adding independent variables and random effects. Nevertheless, starting with
a maximal, valid model is important for ensuring that future revisions are also valid (\dcValidity).





\subsubsection{Deriving Candidate Family and Link Functions} \label{sec:family_link_functions}
The DV's data type determines the set of candidate family and
link functions. For example, numeric 
variables cannot have
binomial or multinomial distributions. Similarly, nominal variables are not
allowed to have Gaussian distributions. Furthermore, each family has a set of
possible link functions. For example, a Gaussian family distribution may have an
Identity, Log, or Square Root link function. The statistics literature documents
 possible combinations of family and link functions for specific data
types~\cite{nelder1972generalized}.

Tisane includes common family distributions as candidate families and their
applicable link functions. In its current implementation, Tisane relies on
\statsmodels~\cite{statsmodelsPaper} for GLMs and
\texttt{pymer4}~\cite{jolly2018pymer4} for GLMMs. As such,
Tisane is limited to the family and link function pairings implemented in these libraries. As
\statsmodels' and \pymer's support for GLMs grows in the future, Tisane can be extended.

\subsection{Eliciting Analyst Input for \Disambiguation}\label{sec:disambiguation}
The disambiguation process provides an opportunity for analysts to explore the
space of generated models based on their original query. Given our design
considerations to prioritize conceptual knowledge (\dcConceptualKnowledge) and
give analysts guidance (\dcGuidance), we designed a GUI to scaffold analysts' reasoning and elicit their input.
For versatility, we implemented Tisane's GUI using Plotly
Dash~\cite{plotlyDash}. Analysts can either execute their Tisane programs and use the
GUI inside a Jupyter notebook (no additional widgets needed) or run
their Tisane programs in an IDE or terminal, in which case Tisane will open the
GUI in a web browser. Figure~\ref{fig:groupExerciseDisambiguation}
gives an overview of the GUI.

Candidate statistical models are organized according to (i) independent
variables  (main effects and interaction effects), (ii) data clustering (random
effects), and (iii) data distribution (family and link functions). In the main
effects tab, Tisane asks analysts if they would like to include additional or
substitute main effects that Tisane infers to be conceptually relevant. In the
interaction effects tab, Tisane suggests moderating relationships to include but
does not automatically include them because analysts may not have specific
hypotheses involving interactions (\dcGuidance). If analysts do not specify any
moderating relationships, Tisane does not suggest any interaction effects,
preventing analysts from including arbitrary interactions that may be
conceptually unfounded (\dcConceptualKnowledge, \dcValidity).

In the data clustering tab, Tisane shows analysts which random effects it
automatically includes based on the selected main and interaction effects. Unlike
main and interaction effects, Tisane automatically includes random effects
in order to maximize model generalizability (\dcValidity). If there is a random
slope and random intercept pertaining to the same unit, Tisane asks analysts if
they should be correlated or uncorrelated. We provide this option because
analysts may have relevant domain expertise to make this decision (\dcGuidance).
By default, Tisane correlates the random slope and random intercept.

The final tab, data distribution, helps analysts examine their data and select
an initial family and link function to try. Appropriate selection of family and
link functions depends on the data type of the dependent variable and the
distribution of model residuals. Therefore, the selection can only be assessed
after choosing a family and link function in the first place.

For an initial statistical model to consider, Tisane narrows the set of family
functions considered based on the declared data type of variables
(see~\autoref{sec:family_link_functions}) and lightweight viability checks, such
as ensuring that a Poisson distribution is only applicable for variables that
have nonnegative integer values. \diff{Tisane asks questions designed to uncover
more semantically meaningful data types (e.g., counts) than are provided at
variable declaration. Analysts without data can answer these questions as they
are planning their studies (\dcStatisticalPlanning). For the selected family
candidate, Tisane automatically selects the default link function based on the
defaults for \statsmodels~\cite{statsmodelsRef} and
\texttt{pymer4}~\cite{jolly2018pymer4}. Analysts can then choose a different
link function, as long as it is supported\footnote{See supplemental material for
a complete listing of Tisane's supported family and link function pairings.}.}

\subsection{Output}
\diff{There are two outputs to interactive compilation: (i) a log of GUI
choices and (ii) an executable modeling script. To increase transparency of the
authoring process, Tisane provides a log of user selections in the GUI as
documentation, which the analyst can include in pre-registrations, for example
(\dcStatisticalPlanning).} In the output script, Tisane includes code to fit the
model and plot residuals against fitted values in order to assess the
appropriateness of family and link functions, as is typical when examining
family and link functions. The output script also includes a comment explaining
what to look for in the plots and an online resource for further reading. Should
analysts revise their choice of family and link functions, they can re-generate
a script through the Tisane GUI.

\section{Case studies with researchers} \label{sec:eval_with_users}

Given Tisane's novel focus on deriving and
guiding analysts toward valid statistical models, we assessed how Tisane affects
data analysis practices in three case studies with researchers. The following research questions guided
the evaluation:
\begin{itemize}
    \item \rqWorkflow How does Tisane's programming and interaction model
    affect how analysts author models? Specifically, what does Tisane make
    noticeably easier or more difficult when conducting an analysis?
    \item \rqCognitive Where do researchers report spending more
    time or attention when using Tisane? How does this compare to their
    fixation during analyses typically?
    \item \rqFuture When do researchers imagine using Tisane
    in future projects, if at all? What additional support do researchers want
    from Tisane? 
\end{itemize}


We recruited researchers through internal message boards and individual
contacts. We intentionally recruited researchers at different stages of the
research process---study planning, data analysis for publication, and ongoing
model building and maintenance. We believed this could help us more holistically
evaluate Tisane's impact on data analysis. \diff{We met with researchers over Zoom
(R1, R3) and in person (R2) to discuss their use cases, observe them use
Tisane for the first time, and ask for open-ended feedback. We pointed researchers to the Tisane tutorial for
installation instructions and examples but otherwise encouraged the researchers
to work independently. We answered any questions researchers had while using Tisane.}
Each study session lasted approximately 2 hours. At the end, two of the three
researchers (R1, R3) said they planned to use Tisane again over the next two months.


\subsection{Case Study 1: Planning a new study}

R1, a clinical psychology PhD student, had recently submitted a paper and was
planning a follow-up.
R1 reported that she had never taken a formal class on modeling techniques but
taught herself for her last paper. \diff{Her general workflow involved consulting with and mirroring what others in her research group did even if she did not completely understand why.} R1 did not program often but said she
had ``enough coding experience to understand this kind of...[sample
program].'' Although familiar with Python, R1 preferred M+~\cite{mplus} and
SPSS~\cite{spss}. She was interested in using Tisane to brainstorm new
studies and research questions.



\textit{Using Tisane.} After installation, R1 read through one of the
computational notebook examples available in the Tisane GitHub repository.
While reading, R1 asked clarifying questions about the variable types and
syntax. R1 explained that the \texttt{Design} class felt novel because she had
never seen the concept of a study design in data analysis code before. When the
first two authors explained that it was supposed to be the equivalent of the
statement of a study design in a paper, R1 remarked that usually, she ``[kept]
that in [her] head, which [she] probably shouldn't'' (\rqCognitive). Without
a concrete data set, R1 preferred to walk through more examples 
rather than author a script of her own.

\diff{While reading an example,} R1 drew a parallel between the tabs in SPSS dialogs for specifying models and
the tabs in the Tisane GUI, noting that SPSS had a tab for control variables.
R1 also wanted the ability to distinguish between ``control
variables'' and other independent variables in the Tisane GUI. R1 explained that this
would map more closely to how psychologists think about analyses.
Future work could incorporate additional language constructs, such as
a new data type for controls, for different groups of users (\rqFuture).

\diff{At the end of the study session,} R1 remarked how Tisane ``fills in a lot of
the...gaps'' in data analysis (\rqWorkflow, \rqCognitive). The first gap R1
discussed was the \emph{programming gap} between scientists and statistical
tools. R1 believed that, for scientists who were not comfortable with
programming, ``they should probably be running less complex models, or first
learn how to code'' even if the complex models would be most appropriate. The
second gap R1 discussed was the \emph{statistical knowledge gap} in tools. R1 explained that
in her experience, R provides support for more complex models but little
guidance for what those models or statistical tests should be, requiring ``top
down assumption[s].'' Thus, to R1, Tisane bridged the gap between tools like
SPSS and R by requiring minimal programming and providing modeling support. Put
another way, Tisane bridged the gulf of execution~\cite{norman2013doet} for R1
that previous tools had not.


\subsection{Case Study 2: Analyzing data for a paper submission}
R2, a computer science PhD student, had conducted a within-subjects study where
47 participants used four versions of an app for one week each (four
weeks total). The motivating research question was how the different app designs
led to 
psychological dissociation. 
Although R2 had expected to collect multiple survey responses for each
participant each day, they only had
aggregate daily self-report measures due to an error in the database management system.
\diff{In the past, R2 reported having extensively explored their data and
consulting others, but for this paper, they had not explored their data prior to
fitting models because they felt more confident in their modeling skills. For
analyses, R2 preferred R but had general Python programming experience.} Prior
to using Tisane, R2 had authored linear mixed effects models in R for their
study. They were interested in using Tisane to check their analyses prior to
submitting their paper to CHI.

\def\numberofinstances{\texttt{number\_of\_instances}\xspace}

\textit{Using Tisane.} 
R2 wrote their scripts by adapting an
example from the Tisane GitHub repository.
As R2 considered which conceptual relationships to add, they reasoned aloud about
if they should state causal or associative relationships between various measures and dissociation (\rqCognitive). 
After some
deliberation, they said, ``I don't feel comfortable [making a causal
statement],'' and instead specified \texttt{associates\_with} relationships.
R1's hesitation to assert causal relationships confirms prior findings that
specifying formal causal graphs is difficult for domain
researchers~\cite{suzuki2020causal,suzuki2018mechanisms,velentgas2013developing} and our design choice
to allow for association edges.
In addition, R2 was initially unsure about how to specify the
\numberofinstances for their measures since their original study design
was unbalanced. 
\diff{After asking for clarification about \numberofinstances,}
R2 declared all the measures with the parameter
\numberofinstances set equal to \texttt{date}.

Next, R2 ran their script 
and used the Tisane GUI in a browser window. Based on Tisane's recommended
family and link functions, R2 realized the models they had previously authored
in R using a Gaussian family were inappropriate. \diff{Due to a bug that we have since
fixed,} Tisane suggested a Poisson family that R2 used to generate a script, but
this was an invalid choice given that not all dependent variable values were
nonnegative integers. R2 explored other family distributions and generated a new
script using an Inverse Gaussian family. When executed, the second output script
issued an error due to the model inference algorithm failing to converge.
R2 made a note to look into this model further on their own.

\diff{Once finished using Tisane,} R2 commented that their analysis with Tisane was more streamlined (\rqWorkflow) in contrast
to their very first paper where they had tried ``every
single kind of model that [they] could'' until finding ``the one that fits best,''
even if it was ``one that no one would have heard of.''
R2 also stated they would be interested in using Tisane earlier
in their analysis process in the future (\rqFuture).
Based on their experience with Tisane, R2 questioned their previously authored linear mixed effects model, and said it was ``unnerving'' to
discover an issue so close to a deadline. At the same time, they expressed, ``if it's incorrect, I should know
before I submit.'' A day after the study, R2 contacted the authors to inform them that they had decided to
update their analyses from linear mixed effects models to generalized linear
mixed effects models. They reported using the Inverse
Gaussian family after visualizing and checking the distribution of residuals
with help from the output Tisane script. The Inverse Gaussian family was
appropriate because their dependent variable's values were all nonnegative and
displayed a slight positive skew.
R2's experience with Tisane suggests that Tisane can help
researchers catch errors and lead them to re-examine their data, assumptions,
and conclusions.

\subsection{Case Study 3: Developing models to inform future models}
Employed on a research team, R3 analyzes health data at the county, state,
and national levels to estimate health expenditure and inform public policy. R3
develops initial models that are used to validate and generate estimates for
larger, more comprehensive models.
Due to the
scale of data and established collaborative workflows, R3 typically works in a
terminal or RStudio through a computing cluster and had very little experience
with Python. Despite working on statistical models every day, R3 described himself as
``not...a great modeler.''
R3 was interested in using Tisane to
determine what variables to include as random effects in a model.

\def\statename{\texttt{state\_name}\xspace}
\def\yearid{\texttt{year\_id}\xspace}
\def\cardinality{\texttt{cardinality}\xspace}

\textit{Using Tisane.} R3 used Tisane in a local Jupyter notebook as well as on
his team's cluster. R3 used the Tisane API overview reference material on GitHub
to start writing his program, which involved copying and pasting the functions
with their type signatures and then modifying them to match his dataset and
incrementally running the program. The most common mistake R3 made while
authoring his Tisane program was to refer to variables using the string names in
the dataset (e.g., \texttt{"year"}) instead of the variable's alias (e.g.,
\yearid), an idiom common in R but not in Python.


\diff{While authoring his Tisane program,} R3 found the \numberofinstances parameter
redundant, especially because his data is always ``square.'' Every \statename\
in his data set had 30 rows of data, corresponding to the \yearid{}s 1990-2019.
This is in contrast to R2, whose study design was unbalanced and resulted in
variable numbers of observations per participant that needed to be aggregated. Based on R3's feedback, we
added functionality to infer \numberofinstances for each unit, which analysts can inspect by
printing the variable.

\diff{While giving open-ended feedback on Tisane,} R3, similar to R1, liked how Tisane helped ``fill
[the] gap in...[his] knowledge'' (\rqCognitive). Given the diversity of models
R3 works with, R3 found Tisane's focus on GLMs and GLMMs a ``little limiting'' and also
wished to make Tisane ``run without...the mouse'' in a script, as is typical in
his workflow (\rqWorkflow). Specifically, R3 described how he and his
collaborators typically want to explore a space of models and run them in
parallel. Nevertheless, R3 foresaw using Tisane in three types of modeling tasks
common in his work: (i) exploratory modeling to determine if there are any
interesting relationships between variables, (ii) authoring and comparing
multiple models for prediction, and (iii) working out the precise model
specification after identifying variables of interest (\rqFuture).



\subsection{System changes and Takeaways}

\diff{We fixed bugs and iterated on Tisane's GUI based on feedback from
researchers. The largest change we made was to the \diff{data distributions} tab. The \diff{data distributions}
tab we tested with researchers visualized the dependent variables
against simulated distributions of family functions and included the results of the Shapiro-Wilk and D'Agostino and Pearson's normality tests. All three researchers
reported becoming more aware of their data due to the visualizations. However, researchers' enthusiasm for the feature made us wary that visualizing the simulated data 
could mislead less careful analysts to believe that family and link functions pertain to variable
distributions rather than the distributions of the model's residuals. 
To avoid
such errors while still helping analysts become more aware of their data, we
removed the simulated visualizations and normality tests and instead provide questions about the semantic nature of the dependent variable
collected, as discussed in~\autoref{sec:disambiguation}.}

Overall, Tisane streamlines the analysis process (\rqWorkflow) in part because
researchers report formalizing their conceptual knowledge into statistical
models more directly (R1, R2). Although Tisane does not eliminate the need for
model revision, Tisane may scope the revisions analysts consider to significant
issues instead of details that may detract from the analysis goals (R2).
Additionally, researchers reported a perceived shift in their attention from
keeping track of and analyzing all possible modeling paths to their research
questions and data assumptions (\rqCognitive) while planning a new study and
analysis (R1) as well as while preparing a research manuscript (R2). Future adoption of
Tisane may depend on the complexity of analyses (\rqFuture) (R3). For instance, Tisane may provide a
streamlined alternative to false starts due to misspecifications for simpler
analyses (R1, R2, R3). For more complex models and studies, Tisane may act more as a
prototyping tool for statistical models, helping researchers start at a
reasonable model that they can then revise (R2, R3).



\section{Discussion} \label{sec:discussion}

In this work, our motivating question was ``\textbf{How might we derive (initial)
statistical models from knowledge about concepts and data collected?}'' This
question presented two challenges: (i) how to elicit the information necessary
to author a GLM/GLMM and (ii) how to computationally infer a valid statistical
model given this information. To address the first challenge, we designed and
developed \textbf{Tisane's \SDSLlong}. To address the second challenge, we developed a
\textbf{graph representation} that Tisane traverses to derive candidate statistical
models. We also developed a novel interaction model that involves \textbf{interactive
compilation} to address both challenges. Throughout the design process,
we employed statistical methods and theory, theories of how people analyze data,
and an iterative design process with researchers. When using Tisane, researchers
in our case studies reported focusing more on their analysis goals and becoming
more aware of their assumptions and even identified and avoided previous
analysis mistakes. \diff{Below, we reflect on future opportunities for Tisane to further enhance statistical practice, interpretation of results, and the end-to-end data analysis pipeline.}


\paragraph{Design for statistical validity.} Campbell's theory of
validity -- encompassing statistical conclusion, internal,
external, and construct
validity~\cite{campbell2015quasiexperimentalDesigns,cook2002generalizedCausalInference}
-- has influenced disciplines widely (e.g.,~\cite{shadish2010campbell}),
including epidemiology (e.g.,~\cite{matthay2020causalDAGEpi}), software
engineering (e.g.,~\cite{neto2013conceptual}), and
psychology (e.g.,~\cite{campbell2015quasiexperimentalDesigns}).
Viewed through the Campbellian framework, Tisane helps analysts avoid
four common threats to statistical conclusion and external
validity: (i) violation of statistical method assumptions, (ii) fishing for
statistical results, (iii) not accounting for the influence of specific
units, and (iv) overlooking the influence of data collection procedures on
outcomes~\cite{cook2002generalizedCausalInference}.

Tisane fills a need to align analysts' conceptual models with the statistical
models they want to implement but find difficult to express with the current
tools available. By integrating conceptual, data, and statistical concerns,
Tisane facilitates the hypothesis
formalization~\cite{jun2021hypothesisFormalization} process, which can be an
error-prone and cognitively demanding process that existing tools do not yet support.

In the future, we plan to develop additional strategies for enhancing the
validity of analyses authored with Tisane. As discussed in
Section~\ref{sec:disambiguation}, our current approach to family and link
functions is only an initial step. We look forward to developing and
comparing multiple strategies for scaffolding the family and link function
selection and revision process. For example, what if the Tisane GUI allowed
analysts to fit multiple models that varied in their family and link functions,
plotted each model's residuals against the predicted values, and gave analysts
visual guides for comparing models? To avoid false discovery rate inflation,
Tisane could partition analysts' data, fit models to only a subset, and output a
script for fitting a selected model using another subset. Although possible for
large datasets, this strategy would encounter limited statistical power for smaller
datasets. Alternatively, what if Tisane calculated Bayes factors for variables
in the
models~\cite{raftery1996approximate,gelfand1994bayesian,czado2006choosing} after
analysts tried multiple family and link combinations? Carefully balancing
statistical rigor and usefulness to domain researchers who may be statistical
non-experts deserves careful consideration.
\diff{\paragraph{Prevent p-hacking.} Tisane generates a space of possible models
from a set of conceptual and data measurement relationships. By querying
Tisane for a model, analysts will only consider a set of models that are
compatible with these relationships. As a result, Tisane helps analysts avoid
unintentional p-hacking. Especially motivated p-hackers could specify
questionable conceptual and data measurement relationships to manipulate the
space of models Tisane generates. However, in this case, review or inspection of
the Tisane program during pre-registration or peer-review, for example, could
identify such malicious practices. In these ways, we believe that p-hacking is
more difficult in Tisane than in existing analysis tools.}

\diff{Future work to further discourage p-hacking could extend Tisane to conduct a
sensitivity analysis on the space of possible models and only report models and
results that are robust across the space. A challenge in this approach is that statistical
non-experts may need more scaffolding to understand and interpret the results of
sensitivity analyses.}

\diff{\paragraph{Scaffold interpretation of statistical results.}
Tisane's focus is on authoring GLMs/GLMMs, but accurate interpretation is also necessary. For instance, analysts may need
help interpreting what their statistical models and results mean in relation to
their input conceptual models. Do the results suggest their conceptual model is
correct? What kind of inferences should they make? Future work should address
these interpretation challenges, which may require eliciting hypotheses and expected results from analysts.}

\diff{Although researchers in our pilot or case studies did not presume
Tisane helped with formal causal analysis, the ability to express causal
relationships (\texttt{causes}) may lead some analysts to erroneously assume
that their models assess causality. Changing the name of the language construct
and/or building out support to interpret GLM/GLMM results may resolve this concern. One way to support accurate interpretation and reporting could be to
output a figure representing the input conceptual model along with visual
summaries of the data and/or statistical model for direct inclusion in publications. 
Tisane could also allow analysts to annotate their disambiguation decisions with
their own rationale and provide a richer log of selections than currently
supported. Tisane could even accept these augmented logs to save the state of the GUI in between analysis sessions.}

\paragraph{Provide discipline-specific language support.} When designing
Tisane's \SDSLlong, we analyzed and developed language constructs common across
existing libraries for study design (see supplemental material). In our case
studies, we found that researchers had different conventions for describing
their data (``unbalanced'' (R2) vs. always ``square'' (R3)) and models (e.g.,
``controls'' (R1) vs. ``covariates'' (R3)). This observation suggests
opportunities to increase the usability of Tisane's \SDSL by providing syntactic
sugar that may be more familiar to users.
In the future, we plan to formally assess usability and identify ``natural''
programming~\cite{myers2004natural} constructs that differ across disciplines.

An additional strategy for supporting more discipline-specific programming
models and analysis needs is to integrate Tisane with existing study design
libraries. For example, HCI researchers may find the lower-level randomization
details that Touchstone2's interface~\cite{eiselmayer2019touchstone2} provides
more natural. A system could summarize these details into the higher-level data
measurement relationships in Tisane to bootstrap interactive compilation and output a possible statistical model. In this way, Tisane's graph IR can provide a ``shared
representation''~\cite{heer2019agency} between study design tools and Tisane.

\paragraph{Integrate into end-to-end analysis workflows.} Researchers in our
case study were more comfortable with R. R1 and R3 expressed it could be helpful
to have Tisane in R as an RStudio plug-in, for example, to fit into their
workflows. As more users adopt Tisane, we will add an implementation in R.

Moreover, analysts may need to add or remove variables from Tisane's generated
statistical models in order to accommodate 
model convergence failures, new data, or changing domain knowledge. However, adding or removing 
variables may subtly change the hypotheses analysts can test statistically. We
look forward to extending Tisane to support model iteration, which presents two
technical challenges: (i) recognizing when conceptual revisions are necessary
and (ii) identifying and suggesting model changes that maintain conceptual
validity or, at the very least, quantify conceptual shifts. Furthermore, in the model revision process, analysts may consider
multiple alternatives. As R3 described, he preferred to run multiple variations
of a model and compare them, a workflow akin to a multiverse
analysis~\cite{steegen2016increasing}. Given that Tisane already generates a
combinatorial space of candidate statistical models, Tisane could generate a
multiverse script for Boba~\cite{liu2020boba} instead. A multiverse could help
check the robustness of findings, and Boba's visual analyzer could help analysts
further develop an understanding of their data and modeling choices. A
multiverse may also help analysts explore and compare family and link
combinations as well.

Tisane is one tool designed to enable analysts with limited statistical expertise to
author valid statistical models. Tisane enables future possibilities and raises
open research questions for creating an ecosystem of analysis tools that align
tool interfaces with analysts' conceptual goals.

\begin{acks} 
We thank Yang Liu and Younghoon Kim for early feedback on Tisane's API; Leilani
Battle, Matthew Conlen, Sherry Wu, and Rock Pang for feedback on early drafts of
this paper; Tyler McCormick for feedback on the project and paper; and Maureen
Daum for insightful conversations about how Tisane's graph IR relates to data
models. We also thank the anonymous reviewers who provided valuable feedback.
\end{acks}

\bibliographystyle{ACM-Reference-Format}
\bibliography{tisane-paper}

\end{document}